\documentclass{article} 
\usepackage{iclr2026_conference,times}

\usepackage{amsmath,amsfonts,bm}

\def\eqref#1{equation~\ref{#1}}

\def\1{\bm{1}}

\DeclareMathAlphabet{\mathsfit}{\encodingdefault}{\sfdefault}{m}{sl}
\SetMathAlphabet{\mathsfit}{bold}{\encodingdefault}{\sfdefault}{bx}{n}

\usepackage{hyperref}
\usepackage{url}
\usepackage{graphicx}
\usepackage{booktabs}
\usepackage{graphicx}
\usepackage{caption}
\usepackage{subcaption}
\usepackage{multirow}
\usepackage{svg}
\usepackage{tabularx}
\usepackage[ruled,vlined]{algorithm2e}
\usepackage{enumitem}
\usepackage[table]{xcolor}
\usepackage{fontawesome5}

\title{SpecExit: Accelerating Large Reasoning Model via Speculative Exit}

\author{
\textbf{Rubing Yang}\thanks{Equal contribution.},
~~\textbf{Huajun Bai}\footnotemark[1],
~~\textbf{Song Liu},
~~\textbf{Guanghua Yu}\thanks{Corresponding author.},
~~\textbf{Runzhi Fan},
~~\textbf{Yanbin Dang},
\\
\textbf{Jiejing Zhang},
~~\textbf{Kai Liu},
~~\textbf{Jianchen Zhu},
~~\textbf{Peng Chen}
\\
\\
\multicolumn{1}{p{12.8cm}}{\centering Tencent} \\
\multicolumn{1}{c}{\texttt{\{rubingyang, huajunbai, lucayu\}@tencent.com}}
}

\begin{document}

\maketitle




\begin{abstract} 
Despite their strong performance on reasoning tasks, large reasoning models (LRMs) often suffer from overthinking, producing unnecessarily long outputs and incurring high end-to-end latency, a significant limitation to their real-world deployment.
To address overthinking, early-exit mechanisms have been proposed to terminate reasoning before typical completion, showing that this approach can effectively shorten generation length with minimal impact on accuracy. However, their reliance on probing mechanisms introduces a detection overhead that limits their end-to-end latency gains and compromises their generalizability across diverse problems.
Inspired by the use of hidden states in speculative decoding, we propose \textbf{SpecExit}, a novel framework that predicts both future tokens and an early-exit signal directly from a lightweight draft model without probing overhead.
Our method offers significant improvements, reducing average generation length by 66\% and achieving a 2.5x speedup in end-to-end latency compared to the speculative decoding baseline, without compromising accuracy.
Our method leverages the inherent signals from hidden states to provide effective early-exit signals, suggesting broader use of hidden states for efficient reasoning.
Our code is available at: \url{https://github.com/Tencent/AngelSlim}.
\end{abstract}



\section{Introduction}
\label{introduction}


Large reasoning models (LRMs) such as OpenAI-o1 \citep{o1}, DeepSeek-R1 \citep{deepseekai2025deepseekr1incentivizingreasoningcapability} and Qwen \citep{qwen2025qwen25technicalreport} have recently achieved state-of-the-art performance in complex tasks. 
These models follow the test-time scaling law \citep{snell2024scaling, brown2024monkeys, muennighoff2025s1}, where generating longer chain-of-thought (CoT) sequences \citep{wei2022chain} generally enhances model performance. 
However, this reliance on extended reasoning often leads to an \textit{overthinking} problem, where models produce unnecessarily verbose outputs.
This redundancy leads to both excessive token usage and high end-to-end latency, which limits LRMs' practical deployment.

To mitigate overthinking, researchers have proposed both inference-time and training-based strategies \citep{sui2025stopoverthinkingsurveyefficient}. 
Inference-time early-exit methods \citep{yang2025dynamic, fu2024efficiently} rely on model-generated signals such as intermediate answers or output logits to terminate decoding once sufficient evidence is detected.
These methods can shorten reasoning length without harming accuracy, but the probing overhead they introduce limits the actual latency gains. 
Training-based approaches, such as reinforcement learning \citep{aggarwal2025l1,yeo2025demystifying} and supervised fine-tuning \citep{ma2025cot,munkhbat2025self}, incur little runtime overhead during inference but risk altering the model’s output distribution. 
As a result, existing methods struggle to deliver consistent improvements in end-to-end efficiency.

Speculative decoding \citep{chen2023accelerating,leviathan2023fast} is a promising approach that improves efficiency without altering the target model’s outputs. A lightweight draft model proposes multiple candidate tokens, which the target model verifies in parallel, substantially reducing per-token latency. However, this strategy alone does not resolve the overthinking problem, as models still generate the full CoT. 
Recent advanced speculative decoding methods \citep{li2024eagle, zhang_learning_2025} exploit hidden states to predict several future tokens. Other studies \citep{lin_controlling_2025,dong_emergent_2025} also show that hidden states encode richer predictive signals beyond next-token probabilities. 
\begin{figure*}[!htbp]
    \centering
    \begin{subfigure}[t]{0.33\textwidth}
        \centering
        \includegraphics[width=\textwidth]{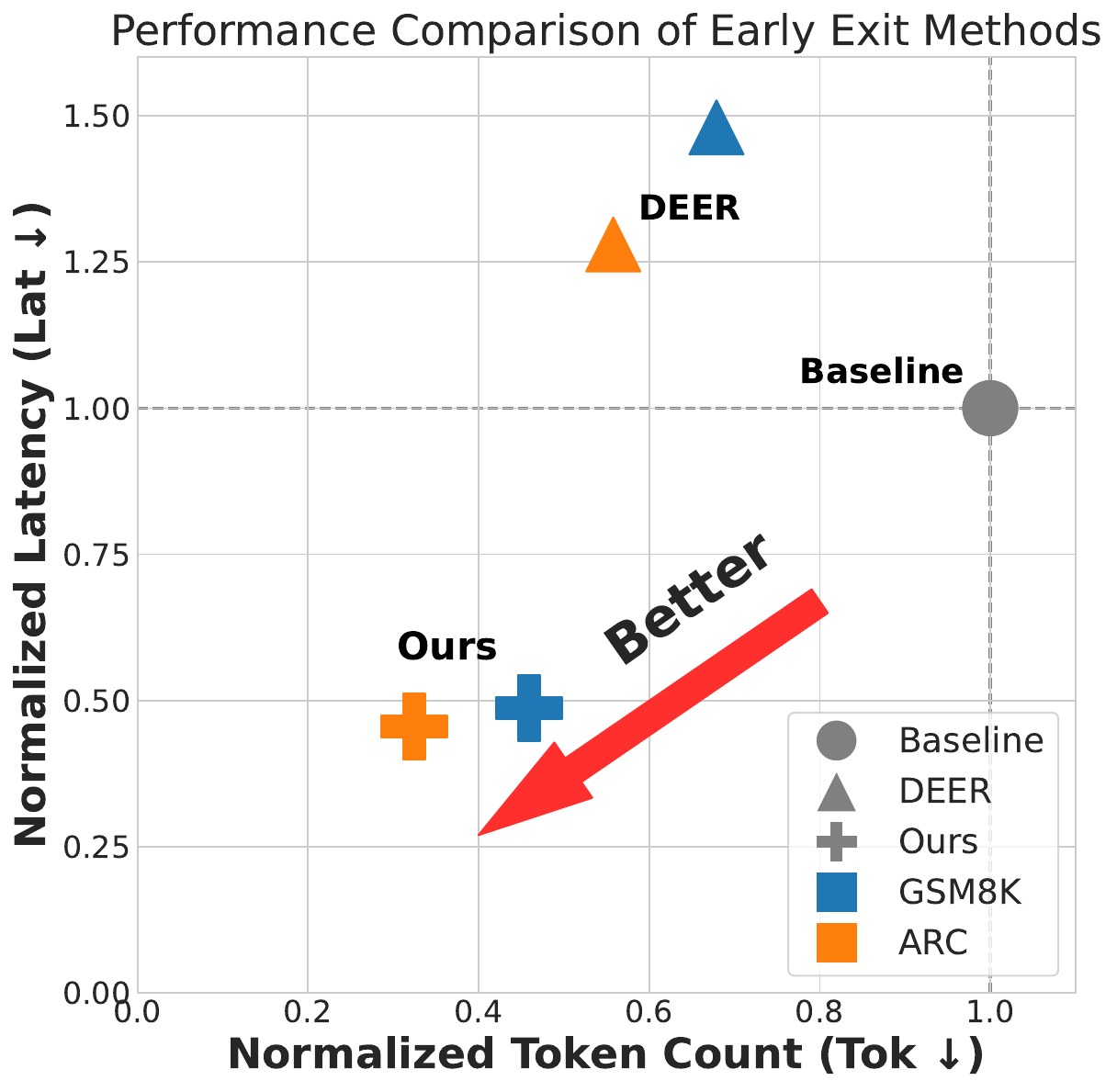}
        \caption{Efficiency Comparison}
        \label{fig:fig1a_metrics}
    \end{subfigure}
    \hfill
    \begin{subfigure}[t]{0.64\textwidth}
        \centering
        \includegraphics[width=\textwidth]{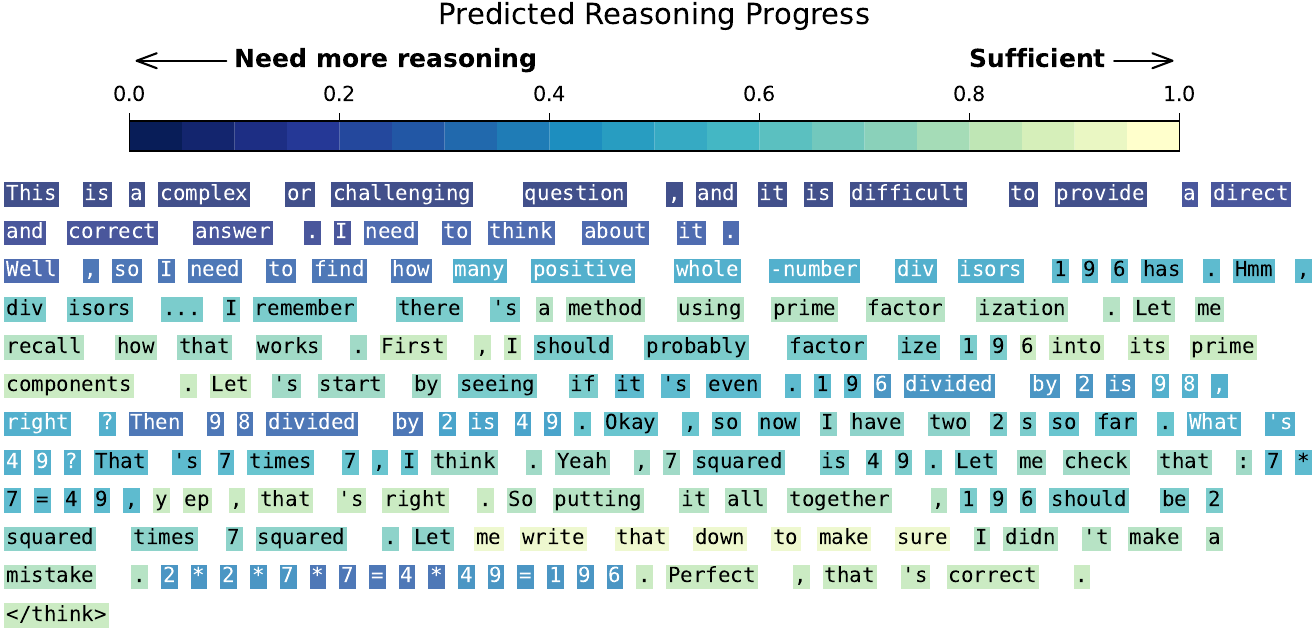}
        \caption{Predicted Reasoning Progress.}
        \label{fig:fig1b_hidden_states}
    \end{subfigure}
    \caption{Effectiveness of the proposed method. (a) Statistical comparison showing that our approach produces shorter reasoning chains and faster inference than baselines. (b) visualizes the predicted reasoning progress on a MATH500 example, where darker colors denote insufficient reasoning and lighter colors denote sufficiency, demonstrating valuable signals can be extracted from hidden states regarding the model's reasoning process.}
    \label{fig:logits_failure_coding}
\end{figure*}

In this work, we introduce \textbf{SpecExit}, a reasoning-aware early-exit framework that leverages draft model hidden states not only to anticipate future tokens but also to predict early-exit signals. Unlike prior probing-based approaches, SpecExit requires no modifications to the target model and incurs no additional detection overhead. Instead, it extends the lightweight draft model with auxiliary prediction heads, enabling it to jointly output token distributions and reasoning-related signals in a single forward pass. By exploiting the latent information embedded in hidden states, SpecExit provides reliable criteria for dynamically terminating chain-of-thought generation when sufficient reasoning has been achieved.

We validate SpecExit on state-of-the-art reasoning models across mathematical, scientific, and logical benchmarks. For Qwen3-4B-Thinking-2507 and DeepSeek-R1-Distill-Llama-8B models, SpecExit shortens generation length by 66\%, and speeds up end-to-end latency by up to 2.5x compared with the speculative decoding baseline. 
Our contributions can be summarized as follows:
\begin{itemize}[leftmargin=*,itemsep=0.12em]
  \item \textbf{Signals Extracted for Early Exit}. We derive early-exit signals from hidden features and integrate them into speculative decoding, enabling reliable early exit for efficient reasoning. 
  \item \textbf{General and Practical Framework}. We implement SpecExit, a reasoning-aware early-exit framework, in both PyTorch and vLLM, making it easy to deploy across diverse inference environments. 
  \item \textbf{Substantial End-to-End Performance Gains}. SpecExit reduces reasoning length by 66\% and achieves up to 2.5$\times$ lower latency than speculative decoding while maintaining accuracy.
\end{itemize}
\vspace{-1.0em}
\section{Motivation}
Current early-exit methods face two challenges: runtime overhead from probing and limited generality from task-specific prompts. 
Since hidden states already reflect reasoning sufficiency, we propose leveraging them to replace costly probing for faster and more reliable inference across diverse tasks.
\begin{figure}[h]
\begin{center}
\includegraphics[width=0.99\linewidth]{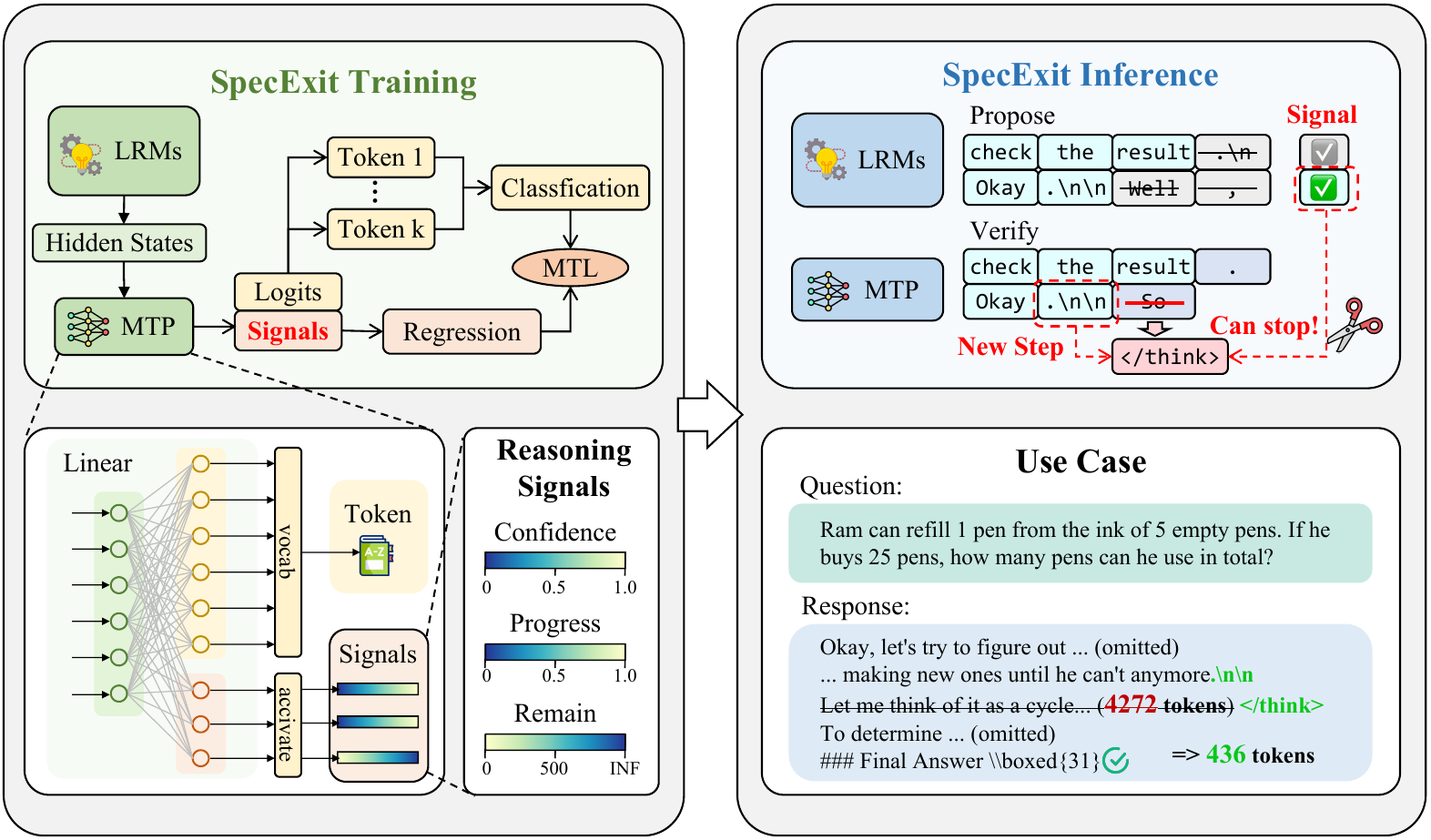}
\end{center}
\caption{Overall architecture of the proposed SpecExit framework. The Multi-Token Prediction (MTP) layer is augmented to output both token logits and auxiliary signals. Training is performed with Multi-Task Learning (MTL), while at inference these signals guide speculative early stopping without modifying the backbone model. The example illustrates how redundant reasoning steps can be pruned while preserving final answer quality.}
\label{fig:method_overall}
\end{figure}

\vspace{-1.3em} 
\textbf{Probing Overhead and Limited Generalizability.} 
Probing-based early-exit methods suffer from both latency overhead and poor generalizability across tasks. Shortened reasoning trajectories do not effectively translate into latency improvements. As shown in Figure~\ref{fig:fig1a_metrics}, 
DEER \citep{yang2025dynamic} reduces the generation length of GSM8K \citep{cobbe2021training} and ARC-Challenge\citep{clark2018think} by 32\% and 44\% respectively on Qwen3-4B-Thinking-2507, but even increases end-to-end latency than vanilla baseline.
The gap arises because probing introduces extra computation, and its effectiveness is highly sensitive to both the task and the model. A probing phrase like ``Final Answer is'' elicits effective intermediate answers in math problems, but fails in coding tasks where extra tokens are still generated. This drawback not only undermines generalizability across domains but also limits true latency savings. 

\textbf{Signals from Hidden States}. 
Our preliminary experiments show that models' hidden states encode informative signals about its reasoning process.
As illustrated in Figure~\ref{fig:fig1b_hidden_states}, we use a simple Multi-Layer Perceptron (MLP) trained on hidden states to predict reasoning progress. For complex tasks, the predicted signals appear darker at the beginning, reflecting the need for continued reasoning, but gradually shift to lighter colors as the model approaches a sufficient chain of thought. This progression suggests that hidden states provide fine-grained indicators of task complexity and reasoning sufficiency. Leveraging these internal signals offers an efficient alternative to costly probing, motivating our approach of utilizing signals from hidden states.

\section{Method}


\subsection{Overall Architecture}

\textbf{SpecExit Framework}. The overall design of this work aims to incorporate additional learnable signals into large model reasoning, thereby providing explicit decision-making criteria for early stopping. In the decoding process of large language models, hidden states not only encode semantic and contextual information for next-token prediction but also implicitly contain higher-level cues related to reasoning progress, generation quality, and content completeness. The Multi-Token Prediction (MTP) mechanism leverages these hidden states to project into the vocabulary space and predict multiple future tokens simultaneously, thereby improving inference efficiency. Inspired by this, we extend the MTP layer while keeping the backbone language model unchanged, introducing auxiliary prediction heads that allow the model to explicitly generate reasoning-related signals, including confidence, reasoning progress, and remaining reasoning length, alongside token distributions. This design preserves the original language modeling ability while providing learnable auxiliary variables for inference control, enabling efficient and dynamic reasoning regulation. The overall architecture of the SpecExit framework is shown in Figure~\ref{fig:method_overall}.

\textbf{Model Structure}. We propose a speculative sampling-based early stopping mechanism for reasoning. In the MTP layer of the model, we extend the linear projection to include additional dimensions. In standard decoding, the hidden state $h \in \mathbb{R}^{D}$ is projected into the vocabulary space to predict the next token distribution, where $D$ is the hidden size of the model. In our method, the linear layer output is extended as:

\begin{equation}
W h = \big[ W_{\text{tok}} h,\; W_{\text{conf}} h,\; W_{\text{prog}} h,\; W_{\text{rem}} h \big],
\label{eq:wh}
\end{equation}

where $W_{\text{tok}} h$ produces standard token predictions, while $W_{\text{conf}} h, W_{\text{prog}} h, W_{\text{rem}} h$ predict confidence, reasoning progress, and remaining reasoning length, respectively. These additional signals serve as observable indicators for deciding whether to stop reasoning early, thus reducing redundant computation without compromising output quality.


\subsection{Signal-Extracted Training}

\begin{figure}[t]
    \centering
    \begin{minipage}[t]{0.5\linewidth} 
        \centering
        \includegraphics[width=\linewidth]{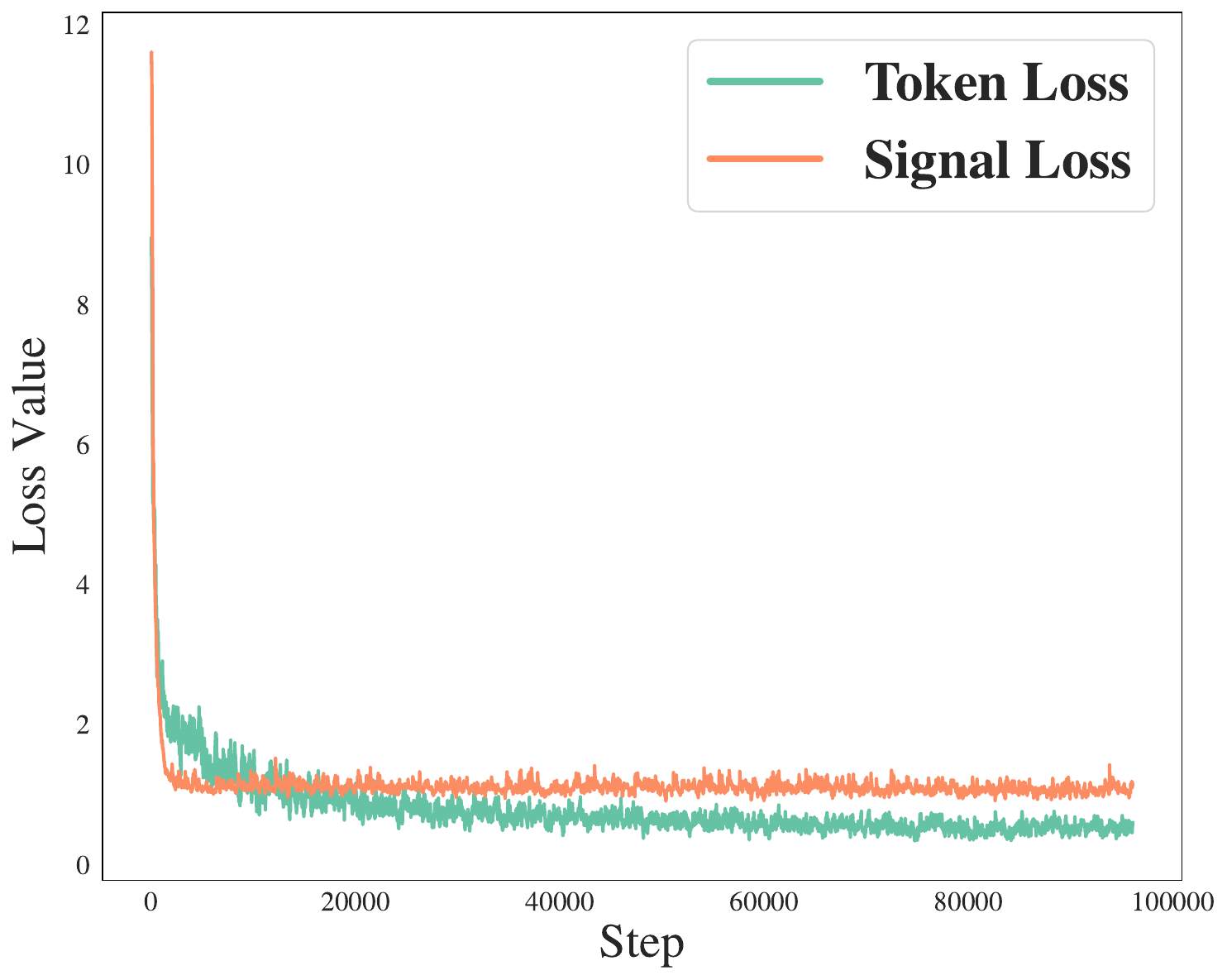}
        \caption{Convergence of token classification loss and signal regression loss during training.}
        \label{fig:loss}
    \end{minipage}\hfill
    \begin{minipage}[t]{0.48\linewidth} 
        \centering
        \includegraphics[width=\linewidth]{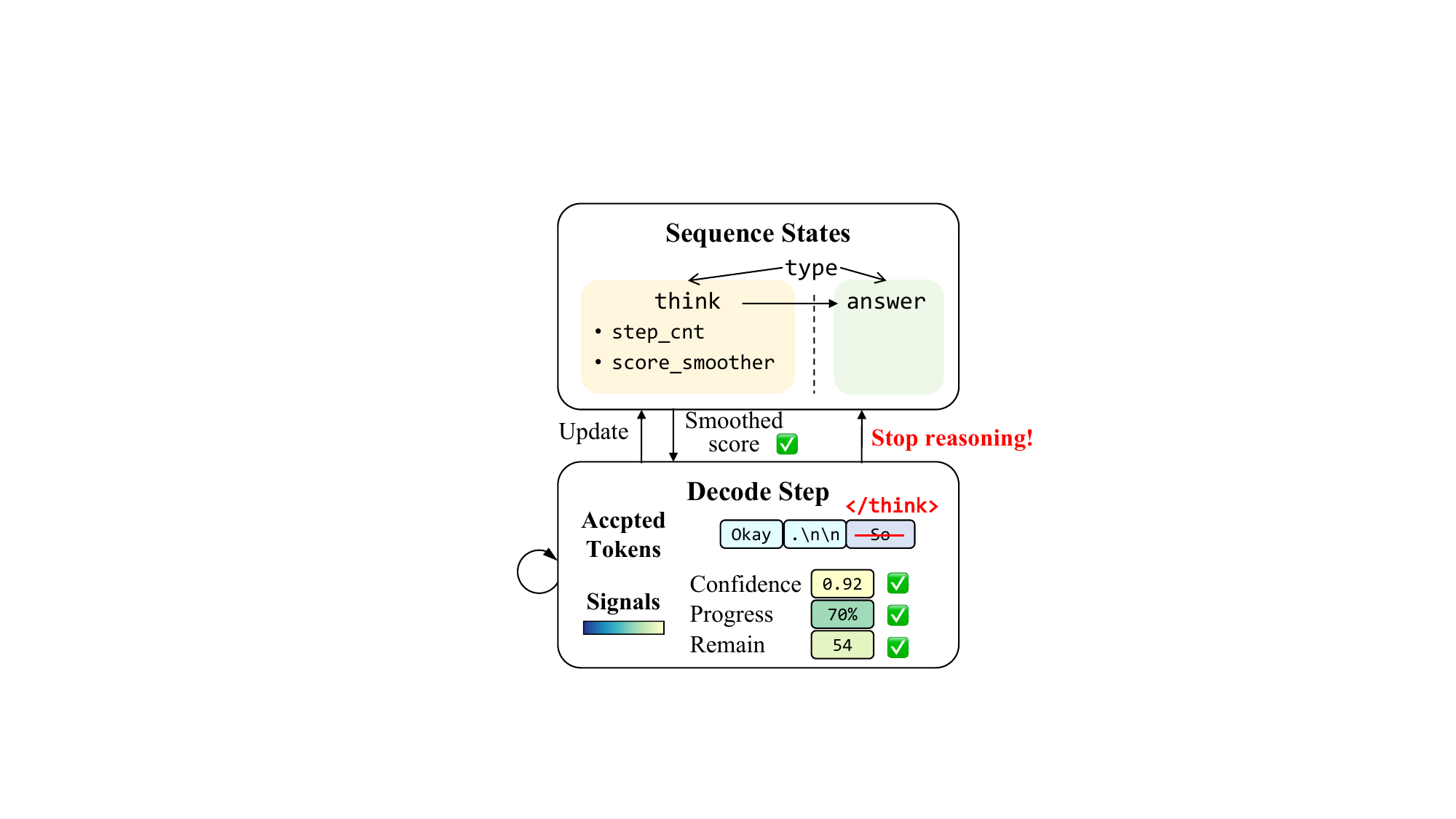}
        \caption{Inference process with signal-guided speculative exit.}
        \label{fig:infer}
    \end{minipage}
\end{figure}

\textbf{Data Construction}. We first obtain the complete response generated by the base language model and extract the reasoning content enclosed within the \verb|<think>| and \verb|</think>| tokens. To identify the effective reasoning trajectory, we iteratively attempt to insert the closing marker \verb|</think>| after each paragraph and verify whether the resulting final answer matches the original output. If the answer remains consistent, the subsequent reasoning content is regarded as redundant.
Consequently, only the minimal reasoning segment required to produce the correct answer is retained as training data.

\textbf{Signal Annotation}. \textsc{Confidence} is defined as the geometric mean of the logit probabilities across prediction steps, reflecting the reliability of the generation; \textsc{remaining} reasoning length is defined as the number of tokens from the initial \verb|<think>| marker to the earliest valid insertion point of \verb|</think>| that still yields the correct answer; \textsc{progress} is represented as a normalized value increasing from 0 to 1, capturing the relative progression of the reasoning CoT.

\textbf{Signal Regression}. We propose a cost-efficient extension by introducing a small number of additional dimensions into the linear projection layer of the MTP module for regressing reasoning signals. These dimensions are orthogonal to the vocabulary classification weights, ensuring that signal regression does not interfere with the convergence of speculative decoding training.

The Multi-Task Learning (MTL) overall training objective jointly optimizes token classification and signal regression, defined as:


\begin{equation}
\mathcal{L} = \mathcal{L}_{\text{cls}} + \lambda_c \mathcal{L}_{\text{conf}} + \lambda_p \mathcal{L}_{\text{prog}} + \lambda_r \mathcal{L}_{\text{rem}},
\label{eq:loss}
\end{equation}

where $\mathcal{L}_{\text{cls}}$ is the standard cross-entropy loss for vocabulary prediction, and $\mathcal{L}_{\text{conf}}, \mathcal{L}_{\text{prog}}, \mathcal{L}_{\text{rem}}$ correspond to the regression losses for confidence, progress, and remaining reasoning length, with $\lambda_c, \lambda_p, \lambda_r$ denoting dynamic weighting coefficients. Specifically:

Confidence and progress are optimized using mean squared error (MSE), remaining reasoning length is optimized with mean squared logarithmic error (MSLE):



\begin{equation}
\mathcal{L}_{\text{conf}} = \frac{1}{N} \sum_{i=1}^{N} (\text{sigmoid}(\hat{c}_i) - c_i)^2, 
\end{equation}
\begin{equation}
\mathcal{L}_{\text{prog}} = \frac{1}{N} \sum_{i=1}^{N} (\text{sigmoid}(\hat{p}_i) - p_i)^2,
\end{equation}
\begin{equation}
\mathcal{L}_{\text{rem}} = \frac{1}{N} \sum_{i=1}^{N} \left(\hat{r}_i - \log(1 + r_i)\right)^2,
\end{equation}

where $\hat{c}_i$, $\hat{p}_i$, and $\hat{r}_i$ represent the model-predicted confidence, progress, and remaining reasoning length, respectively; $c_i$, $p_i$, and $r_i$ denote their corresponding ground-truth values; and $N$ is the total number of samples.

\textbf{Dynamic Weighting}. Since the regression losses of signals converge faster than the token classification loss, we adopt a gradient-based dynamic weighting strategy to balance the contributions of different tasks.  This mechanism assigns higher weights to tasks with smaller gradient magnitudes, preventing tasks with larger gradients from dominating the learning process and ensuring all tasks are effectively optimized. Formally, the mechanism is defined as:


\begin{equation}
\lambda_j = \frac{\|\nabla_\theta \mathcal{L}_j\|}{\sum_{k} \|\nabla_\theta \mathcal{L}_k\|}, \quad
\mathcal{L}_{\text{total}} = \mathcal{L}_{\text{cls}} + \sum_j \lambda_j \mathcal{L}_j,
\label{eq:grad}
\end{equation}

where $\mathcal{L}_j \in \{\mathcal{L}_{\text{conf}}, \mathcal{L}_{\text{prog}}, \mathcal{L}_{\text{rem}}\}$, $\nabla_\theta \mathcal{L}_j$ denotes the gradient of task $j$ with respect to the model parameters, $\mathcal{L}_{\text{cls}}$ is the cross-entropy loss for token classification, and $\lambda_j$ is the dynamically computed weight. This formulation ensures that the gradient contributions of all tasks are balanced, which facilitates stable convergence in multi-task optimization, as shown in Figure~\ref{fig:loss}.






\subsection{Signal-Guided Inference}
\textbf{Overall Procedure}. We build upon the speculative decoding framework, where a smaller draft model first proposes a sequence of candidate tokens, which are then verified in parallel by a larger target model. To evaluate feasibility and efficiency, the inference procedure is implemented on both \textbf{PyTorch} and \textbf{vLLM} frameworks. The central modification lies in the forward pass of the target model. Beyond computing the logits for the next token, we additionally extract the final hidden state corresponding to the last accepted token. This representation is processed through a lightweight linear layer to generate three signals: a confidence score, a progress indicator, and an estimate of the remaining reasoning length, as shown in Figure~\ref{fig:infer}.

\textbf{Stopping Conditions}. To ensure that early-exit decisions occur at semantically coherent boundaries, we introduce a class of special markers called \textit{step split tokens}, which indicate natural segmentation points in the generated text. 
Specifically, step split tokens can be divided into two categories: \textsc{Paragraph Delimiters} (e.g., \verb|.\n\n|), which mark the end of a paragraph or reasoning unit, and \textsc{Discourse Markers} (e.g., "Wait", "But", or "Therefore"), which often signal semantic transitions or logical shifts during reasoning. Since the segmentation strategy based on \textsc{Paragraph Delimiters} is more general, this strategy is adopted by default in subsequent experiments.
Examples of commonly observed discourse markers in reasoning trajectories are shown in Fig.~\ref{fig:wordcloud} (Appendix). When a sampled token belongs to the above set, the early-exit logic is triggered. If the smoothed signal exceeds the predefined threshold, the system determines that the reasoning process has been sufficiently explored. In this case, the accepted output length is truncated at the position of the step split token, and the target model’s recover token is replaced with a special reasoning-end marker (e.g., \verb|</think>|), thereby ensuring that the termination point lies at a natural boundary while maintaining coherence of the generated text. The complete inference process is summarized in Algorithm~\ref{algorithm:inference}.

\begin{algorithm}[htbp]
\SetAlgoLined
\KwIn{Draft model $M_d$, target model $M_t$, tokenizer, thresholds}
\KwOut{Generated sequence $y$}
Define $is\_thinking \leftarrow$ true\;
Define $step\_split\_tokens \leftarrow$ \{ids of ``\verb|\n\n|'', ``\verb|.\n\n|'', ...\}\;
Define $stop\_think\_token \leftarrow$ id of \verb|</think>|\;
\While{not terminated}{
    Extract hidden state of $t_{acpt}$, generate candidate tokens with $M_d$\;
    Compute $signals$ (confidence, progress, remaining)\;
    Set $signals \leftarrow$ update smoothed scores\;
    Concat last accepted token with draft candidates, 
    forward through $M_t$ with tree attention\;
    Accept tokens $t_{acpt}$, accept length $l_{acpt}$, target model recover token $t_{rec}$ \;
    \If{ $is\_thinking$ \textbf{and} $\text{any}(t_{acpt} \in step\_split\_tokens)$ \textbf{and} signals exceed thresholds}{
        Set $l_{acpt}$ $\leftarrow$ corresponding $step\_split\_token$\ position\;
        Set $t_{rec} \leftarrow stop\_think\_token$\;
        Update KV-cache and hidden states accordingly\;
        Set $is\_thinking \leftarrow$ false\;
    }
}
\caption{Inference procedure with signal-guided speculative exit}
\label{algorithm:inference}
\end{algorithm}

\textbf{Signal Smoothing}. Since raw signals may exhibit significant volatility, relying directly on them risks premature or unstable termination. To enhance robustness, we apply an Exponentially Weighted Moving Average (EWMA) to smooth the signals across steps. At each iteration, the smoothed value is updated as a weighted average of the current raw signal and the previous smoothed value, with the smoothing factor controlling the balance between recent and past observations. A smaller factor emphasizes historical stability, yielding smoother trajectories that are less sensitive to transient noise. This ensures that termination decisions reflect consistent trends rather than isolated fluctuations.

\section{Experiments}
\label{experiments}

\subsection{Experimental Setup}


To evaluate the effectiveness of our SpecExit framework, we conducted a comprehensive set of experiments across multiple domains. Specifically, we used the \textbf{GSM8K} \citep{cobbe2021training}, \textbf{MATH500} \citep{math500hendrycks2021measuringmathematicalproblemsolving} and \textbf{AIME} \citep{aime} datasets for mathematical reasoning, the \textbf{HumanEvalPlus} \citep{liu2023your} dataset for coding, the \textbf{GPQA Diamond} \citep{rein2023gpqagraduatelevelgoogleproofqa} dataset for science, and the \textbf{ARC-Challenge} \citep{clark2018think} dataset for logic. Experiments are conducted on two mainstream LRMs: \textbf{Qwen3-4B-Thinking-2507} \citep{qwen2025qwen25technicalreport} and \textbf{DeepSeek-R1-Distilled-Llama-8B} \citep{deepseekai2025deepseekr1incentivizingreasoningcapability}.

We compare our SpecExit method against several baselines: \textbf{Vanilla}, which represents full generation without any early-exit mechanism; \textbf{NoThink} \citep{ma2025reasoning}, which skips the reasoning phase; \textbf{DEER} \citep{yang2025dynamic}, a dynamic early-exit method; and \textbf{EAGLE3} \citep{li2025eagle3scalinginferenceacceleration}, a speculative decoding baseline, which is trained using the same dataset as SpecExit to ensure a fair comparison.
Our performance analysis is based on three key metrics, as detailed in Table \ref{exp_main_results}: \textbf{Accuracy ($\uparrow$)}, \textbf{Token ($\downarrow$)} count and end-to-end \textbf{Latency ($\downarrow$)}. 
All experimental results are obtained by implementing our early-exit strategy in \texttt{vLLM} ~\citep{woosuk2023vllm}, and running inference on an 8×H20 GPU cluster.

\subsection{Main Results}

We first evaluate the proposed SpecExit against baseline reasoning approaches on mathematical, scientific, coding, and logical benchmarks. As shown in Table~\ref{exp_main_results}, SpecExit consistently achieves substantial reductions in both output length and inference latency while maintaining comparable or even higher accuracy. 

Across benchmarks, SpecExit significantly shortens reasoning trajectories, with up to 54\% and 53\% reduction on GSM8K and ARC-Challenge for Qwen3-4B-Thinking-2507, and up to 66\% and 64\% reduction for DeepSeek-R1-Distill-Llama-8B. The reduced reasoning length corresponds to measurable efficiency improvements: SpecExit achieves a 1.9x latency reduction with Qwen3-4B-Thinking-2507 and up to 2.5x speedup with DeepSeek-R1-Distill-Llama-8B on GSM8K, compared with the speculative decoding baseline EAGLE3. Importantly, these gains come only with marginal accuracy differences, confirming that early termination of redundant reasoning does not harm task performance. By contrast, prior inference-time methods primarily focus on reducing output length, but the latency gains they achieve are relatively modest. In some datasets, the additional computational overhead even leads to slower inference than the standard think mode. Notably, for Qwen3-4B-Thinking-2507, inserting the \texttt{</think>} token in the NoThink baseline still fails to suppress reasoning, resulting in even longer outputs than the original Think mode.



Overall, these results demonstrate that SpecExit achieves a favorable balance between efficiency and accuracy, highlighting the practicality of integrating reasoning-aware early-exit strategies into LRMs inference.


\setlength{\tabcolsep}{2.5pt}
\begin{table*}[t]
\centering
\caption{Performance comparison of various reasoning methods on mathematical, scientific, general, and coding benchmarks. “Acc” denotes accuracy, “Tok” denotes token count, and “Lat” denotes total end-to-end latency. $\uparrow$ indicates that higher values are better, while $\downarrow$ indicates that lower values are better.}
{\renewcommand{\arraystretch}{1.3}
\resizebox{\textwidth}{!}{
\begin{tabular}{@{}lcc>{\columncolor{gray!10}}ccc>{\columncolor{gray!10}}ccc>{\columncolor{gray!10}}ccc>{\columncolor{gray!10}}ccc>{\columncolor{gray!10}}ccc>{\columncolor{gray!10}}ccc>{\columncolor{gray!10}}c@{}}
\toprule
 & \multicolumn{9}{c}{\textbf{Math}} 
 & \multicolumn{3}{c}{\textbf{Coding}} 
 & \multicolumn{3}{c}{\textbf{Science}} 
 & \multicolumn{3}{c}{\textbf{Logic}} 
 \\ \cmidrule(l){2-10} \cmidrule(l){11-13} \cmidrule(l){14-16} \cmidrule(l){17-19}
\textbf{Method} 
 & \multicolumn{3}{c}{\textbf{GSM8K}} & \multicolumn{3}{c}{\textbf{MATH500}} 
 & \multicolumn{3}{c}{\textbf{AIME}} & \multicolumn{3}{c}{\textbf{HUMANEVAL+}} 
 & \multicolumn{3}{c}{\textbf{GPQA-D}} 
 & \multicolumn{3}{c}{\textbf{ARC-Challenge}} 
 & \multicolumn{2}{c}{} \\ 
 & {Acc$\uparrow$} & {Tok$\downarrow$} & {Lat$\downarrow$} 
 & {Acc$\uparrow$} & {Tok$\downarrow$} & {Lat$\downarrow$} 
 & {Acc$\uparrow$} & {Tok$\downarrow$} & {Lat$\downarrow$} 
 & {Acc$\uparrow$} & {Tok$\downarrow$} & {Lat$\downarrow$} 
 & {Acc$\uparrow$} & {Tok$\downarrow$} & {Lat$\downarrow$} 
 & {Acc$\uparrow$} & {Tok$\downarrow$} & {Lat$\downarrow$} 
 \\ \midrule
\multicolumn{21}{l}{\textit{\textbf{Qwen3-4B-Thinking-2507}}} \\
\textit{Think}    & 95.3  & 1414   & 155.6 & 96.6  & 6719   & 530.1 & 86.7  & 19577  & 243.3 & 90.9  & 5079   & 175.3 & 68.7  & 9041   & 325.8 & 95.6  & 1812   & 156.5 \\
\textit{NoThink*}  & 95.2  & 1631   & 204.2 & 96.6  & 6395   & 488.5 & 86.7  & 19816  & 243.2 & 88.4  & 4480   & 131.5 & 67.2  & 8833   & 276.8 & 95.1  & 1889   & 159.8 \\
\textit{DEER}     & 94.3  & 960    & 230.3 & 94.4  & 4893   & 519.6 & 70.0  & 17838  & 218.6 & 86.6  & 4079   & 242.4 & 67.2  & 9053   & 505.2 & 94.6  & 1011   & 200.3 \\
\textit{EAGLE3}    & 94.8  & 1408   & 140.3 & 96.6  & 6670   & 395.7 & 80.0  & 19792  & 187.3 & 87.2  & 5178   & 81.7  & 67.7  & 8975   & 212.2 & 95.7  & 1822   & 164.2 \\
\textit{\textbf{SpecExit}} & \textbf{93.8}  & \textbf{649}    & \textbf{75.8}  & \textbf{96.8}  & \textbf{4777}   & \textbf{367.9} & \textbf{90.0}  & \textbf{17769}  & \textbf{206.1} & \textbf{89.6}  & \textbf{4319}   & \textbf{58.4} & \textbf{68.7}  & \textbf{7011}   & \textbf{137.0} & \textbf{94.5}  & \textbf{588}    & \textbf{71.4}  \\
\midrule
\multicolumn{21}{l}{\textit{\textbf{DeepSeek-R1-Distill-Llama-8B}}} \\
\textit{Vanilla}    & 76.4  & 1008 & 629.4 & 81.8  & 6878  & 857.1 & 36.7  & 22170 & 307.0 & 74.4  & 6287 & 445.5 & 43.6  & 8857 & 574.0 & 49.9  & 1917 & 628.5 \\
\textit{NoThink} & 54.6  & 233  & 22.2  & 55.2  & 1643  & 262.8 & 10.0  & 8744  & 184.1 & 46.3  & 472  & 7.3   & 26.8  & 1200 & 166.6 & 12.6  & 135  & 13.6  \\
\textit{DEER}       & 74.7  & 710  & 484.8 & 80.8  & 3533  & 973.3 & 40.0  & 15619 & 272.3 & 79.3  & 4206 & 269.2 & 40.9  & 8492 & 521.5 & 47.5  & 1029 & 531.3 \\
\textit{EAGLE3}     & 79.3  & 976  & 276.9 & 80.8  & 6172  & 593.6 & 30.0  & 25686 & 228.1 & 78.7  & 5312 & 346.5 & 43.9  & 8749 & 420.1 & 59.2  & 1378 & 496.4 \\
\textbf{\textit{SpecExit}}   & \textbf{75.3}  & \textbf{333}  & \textbf{112.6} & \textbf{80.6}  & \textbf{1968}  & \textbf{348.3} & \textbf{36.7}  & \textbf{8160}  & \textbf{176.0} & \textbf{81.7}  & \textbf{3105} & \textbf{118.1} & \textbf{46.0}  & \textbf{6849} & \textbf{307.5} & \textbf{50.3}  & \textbf{500}  & \textbf{253.7} \\
\bottomrule
\end{tabular}
}}
\label{exp_main_results}
\end{table*}

\subsection{Ablation Study}

\begin{figure*}[htbp]
    \centering
    \begin{subfigure}[t]{0.49\textwidth}
        \centering
        \includegraphics[width=\textwidth]{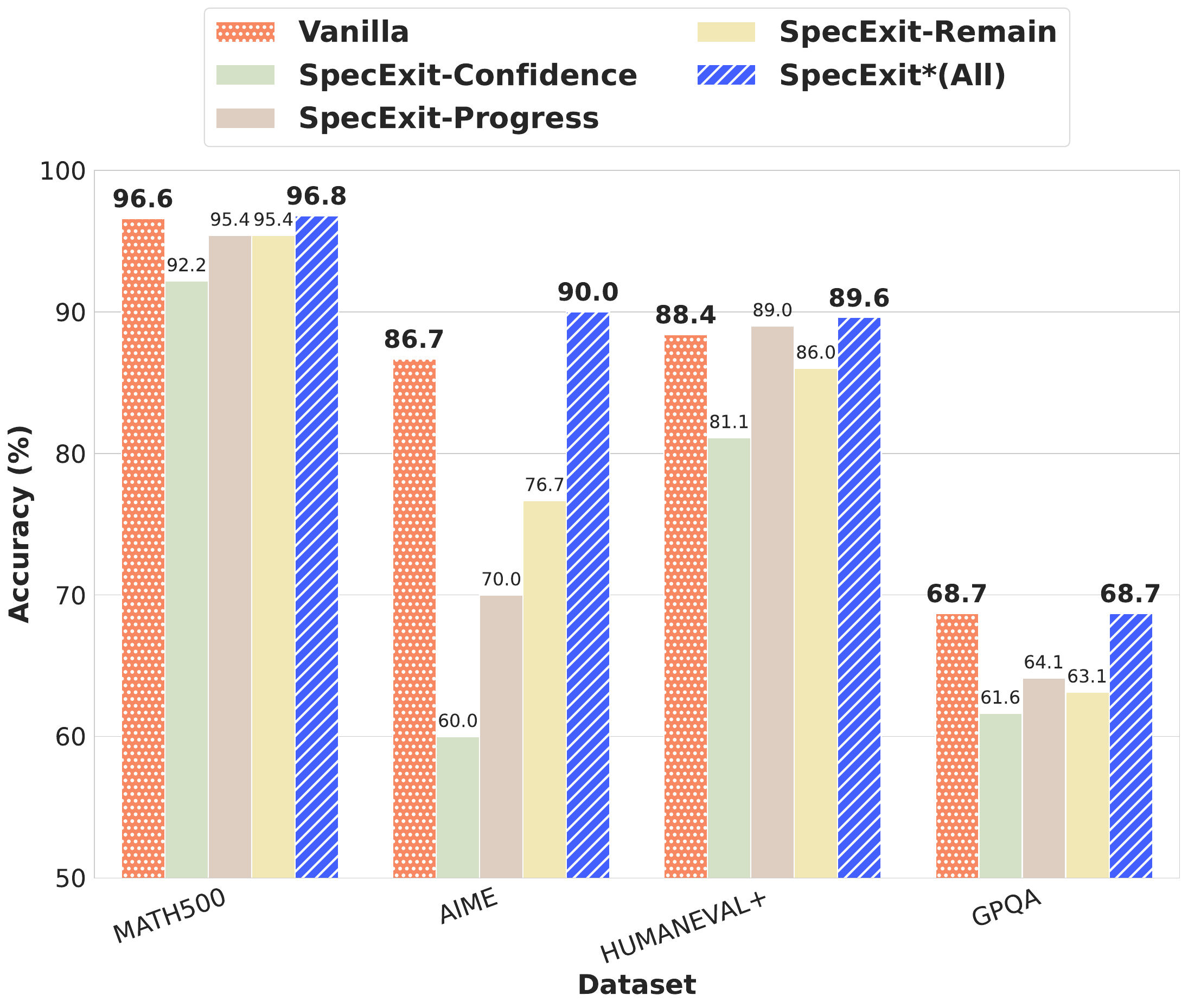}
        \caption{Accuracy comparison.}
        \label{fig:ablation_signal_acc}
    \end{subfigure}
    \hfill
    \begin{subfigure}[t]{0.49\textwidth}
        \centering
        \includegraphics[width=\textwidth]{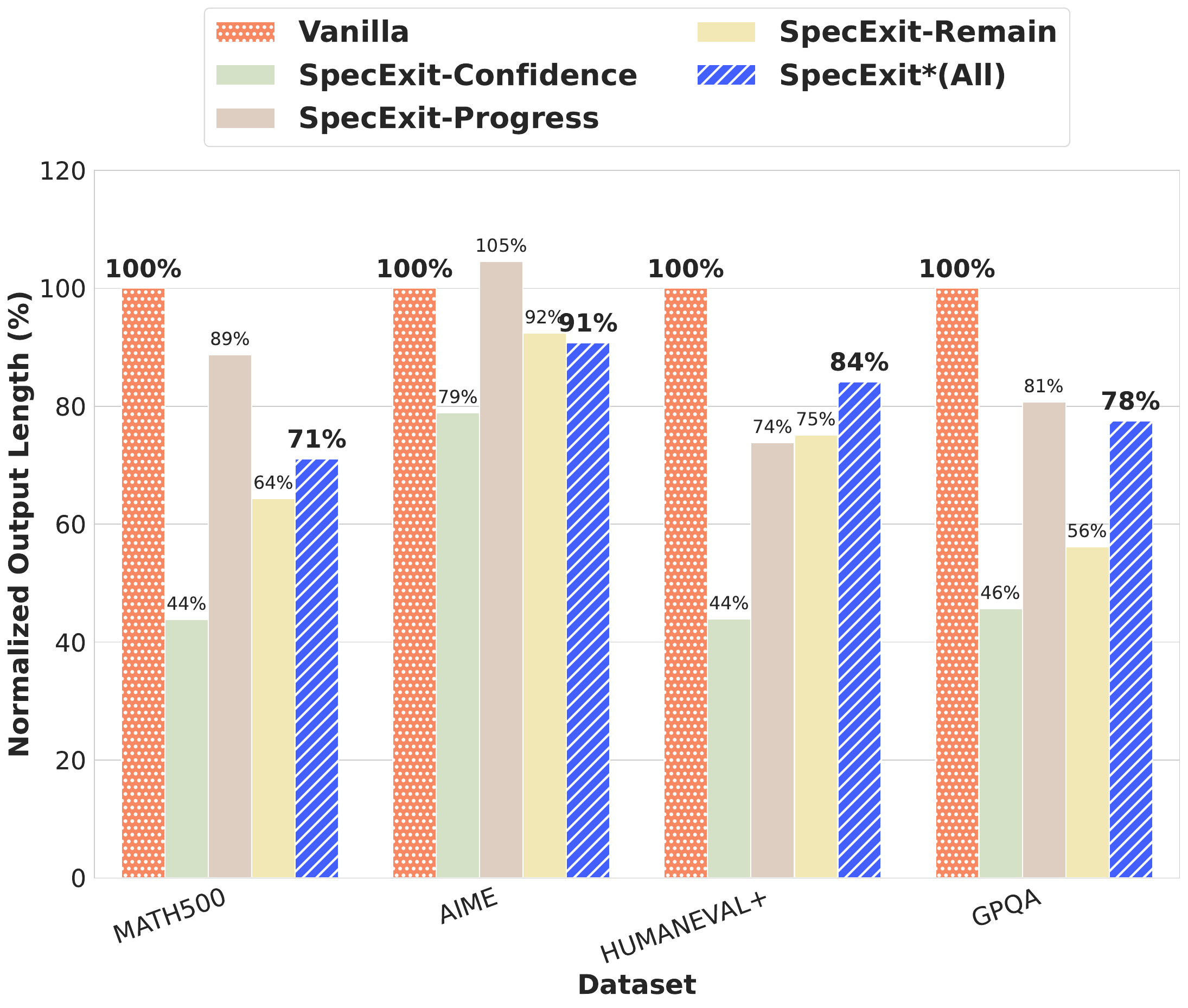}
        \caption{Output length comparison.}
        \label{fig:ablation_signal_tok}
    \end{subfigure}
    \caption{Ablation study of SpecExit signal types on Qwen3-4B-Thinking-2507.}
    \label{fig:ablation_signal}
\end{figure*}

\textbf{Signal Type}. To investigate the impact of individual reasoning signals in SpecExit, we conduct ablation studies on confidence, progress, and remaining token length, along with a combined configuration (SpecExit*) that integrates all three.
As shown in Figure~\ref{fig:ablation_signal}, the confidence-only variant yields the largest token reduction but overestimates the model’s certainty, resulting in noticeable accuracy drops on complex benchmarks.
The predicted reasoning progress increases sharply in the early steps yet continues to fluctuate during iterative reflection. Remaining token length is generally high at the beginning of inference but often triggers premature exits on complicated problems. By integrating all signals, SpecExit* leverages their complementary strengths, preserving competitive accuracy while substantially reducing tokens, demonstrating that multi-signal integration mitigates individual biases and enables more reliable early stopping.

\textbf{Signal Smoothing}. In order to investigate the influence of different smoothing strategies on the stability and performance of early-exit decisions, we conducted a series of ablation experiments comparing multiple approaches. As shown in Table~\ref{tab:ablation_smooth}, removing smoothing increases the volatility of cognitive signals, leading to inconsistent early exits and increased token consumption. The momentum-based prediction strategy significantly reduces token usage, though it may slightly degrade accuracy due to overly aggressive early termination. Smoothing using sliding-window and paragraph-level averaging offers a better trade-off, maintaining accuracy while improving efficiency. Among all methods, Exponential Weighted Moving Average (EWMA) strikes the most consistent balance, providing both stability and reliability. These results demonstrate that appropriate smoothing is essential for reliable early-exit behavior, as it mitigates the influence of transient fluctuations in the raw cognitive signals.


\begin{table*}[htbp]
\caption{Ablation study of different signal smoothing methods on Qwen3-4B-Thinking-2507.}
\centering
\resizebox{0.9\textwidth}{!}{ 
\begin{tabular}{@{}ccccccccccccc@{}}
\toprule
\multirow{2}{*}{Method}            & \multicolumn{2}{c}{\textbf{MATH500}} & \multicolumn{2}{c}{\textbf{AIME}} & \multicolumn{2}{c}{\textbf{HUMANEVAL}+} & \multicolumn{2}{c}{\textbf{GPQA-D}} & \multicolumn{2}{c}{\textbf{Average}} \\ \cmidrule(lr){2-3} \cmidrule(lr){4-5} \cmidrule(lr){6-7} \cmidrule(lr){8-9}  \cmidrule(lr){10-11} 
                                   & {Acc$\uparrow$}               & {Tok$\downarrow$}     & {Acc$\uparrow$}             & {Tok$\downarrow$}    & {Acc$\uparrow$}                 & {Tok$\downarrow$}      & {Acc$\uparrow$}              & {Tok$\downarrow$}   & {Acc$\uparrow$}               & {Tok$\downarrow$}     \\ 
\midrule
\textit{\textbf{Vanilla}}          & 96.60             & 6719    & 86.67           & 19577  & 88.40               & 5133     & 68.69            & 9041  & 85.09             & 10118   \\
\midrule
\textit{\textbf{NoSmooth}}      & 94.20             & 3608    & 73.33           & 17832  & 92.10               & 2789     & 62.12            & 4066  & 80.44             & 7074    \\
\textit{\textbf{Momentum}}          & 91.80             & 2230    & 60.00           & 12427  & 83.50               & 2219     & 64.65            & 3406  & 74.99             & 5071    \\
\textit{\textbf{Sliding Window}}   & 95.20             & 4444    & 80.00           & 19184  & 86.60               & 4342     & 62.12            & 4738  & 80.98             & 8177    \\
\textit{\textbf{Paragraph Mean}}   & 95.40             & 4285    & 76.67           & 18231  & 84.80               & 4569     & 65.66            & 4726  & 80.63             & 7953    \\
\textit{\textbf{SpecExit* (EWMA)}} & \textbf{96.80}    & 4777    & \textbf{90.00}  & 17769  & \textbf{89.60}      & 4319     & \textbf{68.69}   & 7011  & \textbf{86.27}    & 8469    \\ 
\bottomrule
\end{tabular}
}
\label{tab:ablation_smooth}
\end{table*}

\begin{figure*}[htbp]
    \centering
    \begin{subfigure}[t]{0.49\textwidth}
        \centering
        \includegraphics[width=\textwidth]{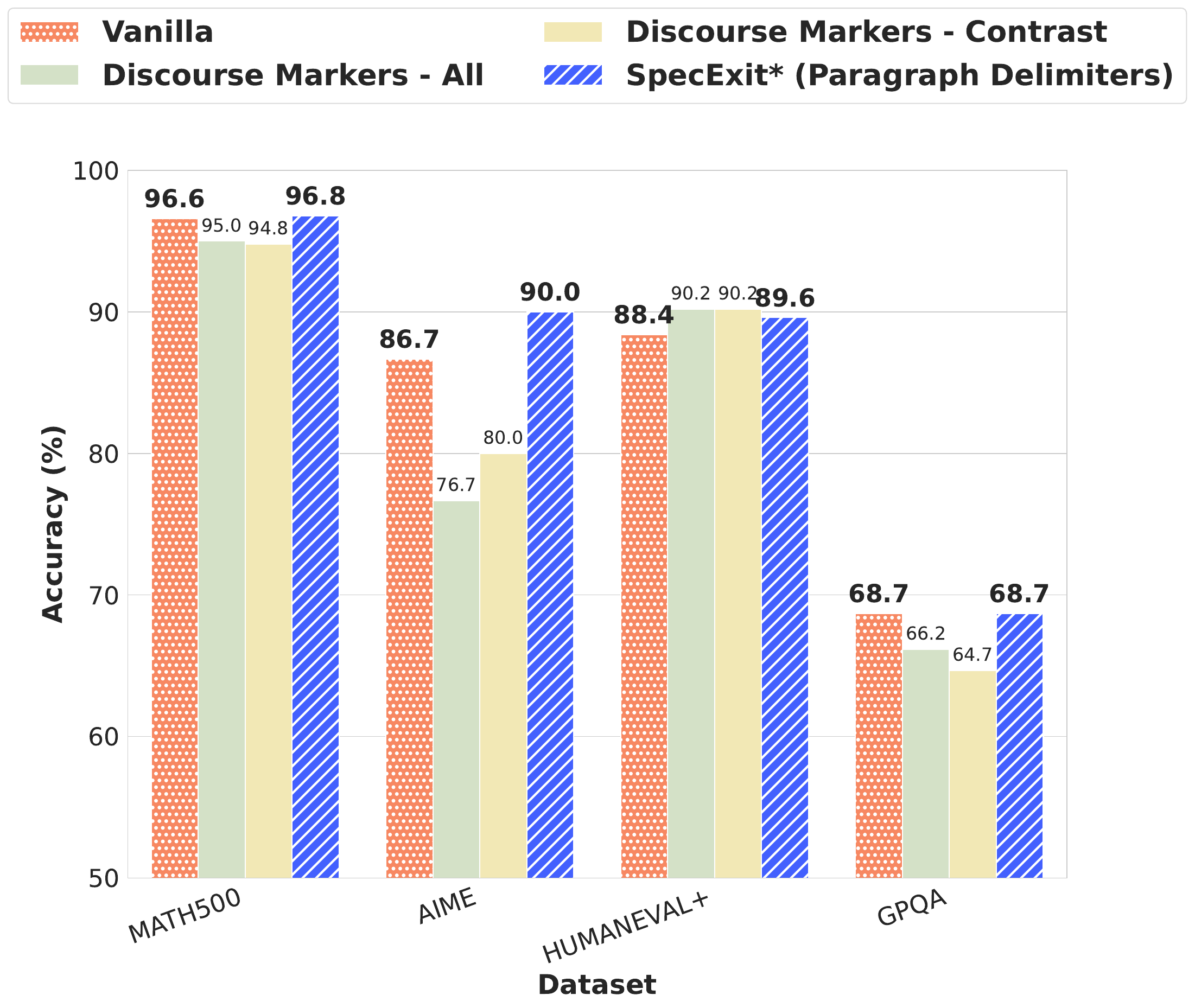}
        \caption{Accuracy comparison.}
        \label{fig:ablation_step_split_tokens_acc}
    \end{subfigure}
    \hfill
    \begin{subfigure}[t]{0.49\textwidth}
        \centering
        \includegraphics[width=\textwidth]{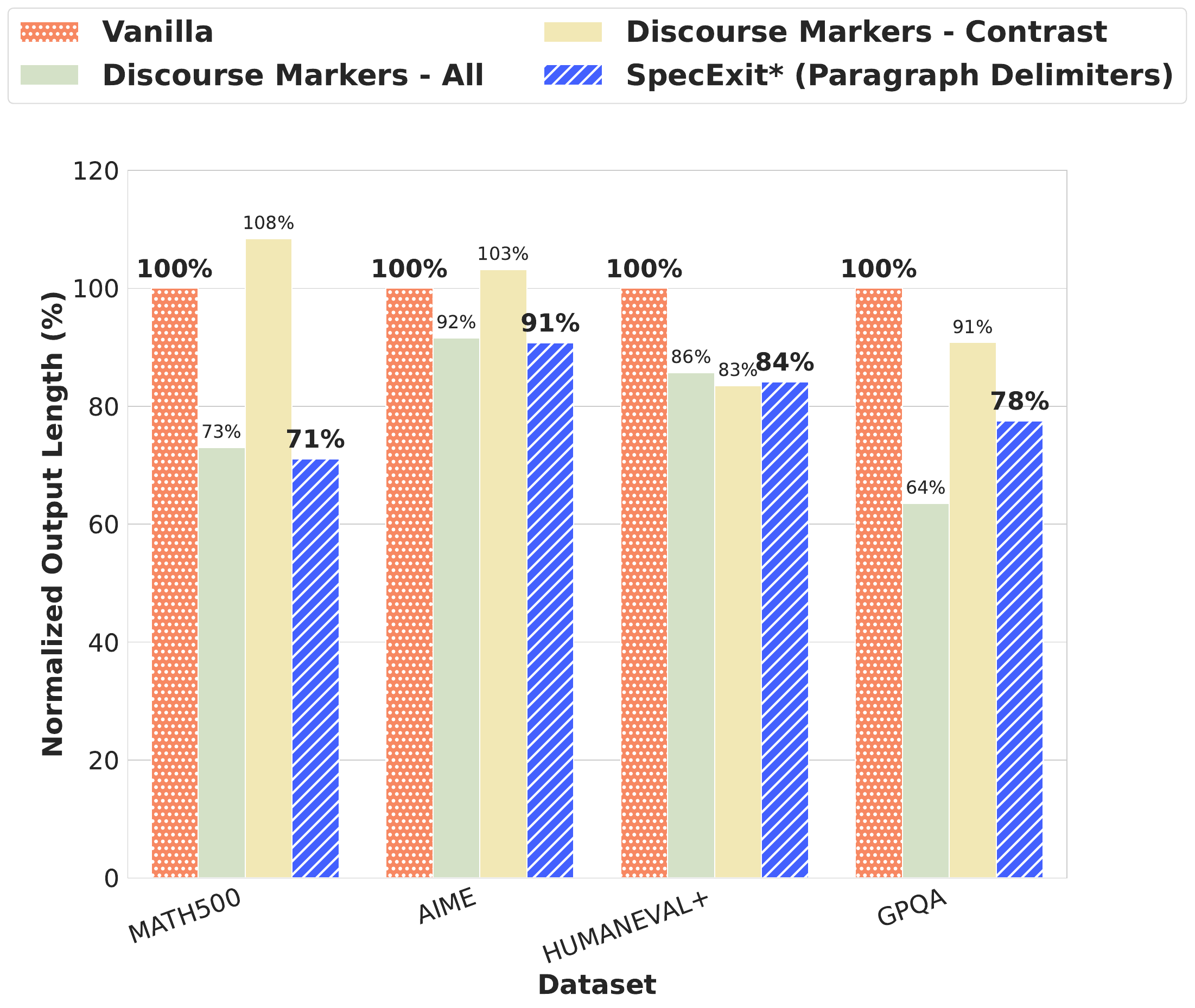}
        \caption{Output length comparison.}
        \label{fig:ablation_step_split_tokens_tok}
    \end{subfigure}
    \caption{Ablation study of step split tokens strategies on Qwen3-4B-Thinking-2507.}
    \label{fig:ablation_step_split_tokens}
\end{figure*}

\textbf{Step Split Tokens}. To evaluate the influence of different step split strategies on early-exit performance, we conducted ablation experiments comparing paragraph delimiters, general discourse markers, and a contrastive subset of discourse markers. Discourse markers indicate semantic transitions or reasoning shifts, but are dependent on the underlying data and model, limiting their generality. 
In prior work on dynamic early-exit methods, contrastive subsets of discourse markers (e.g., “Wait”, “But”, “Alternatively”) are frequently used to capture reasoning-relevant transitions.
In contrast, paragraph delimiters (\verb|\n\n|) provide a more general segmentation that does not rely on model-specific or dataset-specific patterns. As shown in Figure~\ref{fig:ablation_step_split_tokens}, using paragraph delimiters achieves competitive accuracy and token reduction, demonstrating that a general segmentation strategy can be effective for early-exit decisions while maintaining coherence in reasoning trajectories.

In summary, our ablation studies on signal types, smoothing strategies, and step split methods provide key insights for improving early-exit decision-making. The integration of multiple signals strikes the best balance between accuracy and token efficiency, while appropriate smoothing methods stabilize cognitive signals and enhance the consistency of early exits. Additionally, using general segmentation strategies, such as paragraph delimiters, improves the generalizability of early-exit systems across diverse datasets. These findings emphasize the importance of a holistic approach, where complementary strategies jointly enhance both efficiency and reliability.

\section{Related Work} 
\label{related_work}
\textbf{Efficient Reasoning.}
To mitigate unnecessary CoT generation in LRMs \citep{chen2025think23overthinkingo1like,sui2025stopoverthinkingsurveyefficient}, prior work has explored both training-based and inference-time strategies.
Training-based approaches typically modify model behavior through reinforcement learning with length-sensitive objectives \citep{aggarwal2025l1,yeo2025demystifying,shen2025dast} or supervised fine-tuning on reasoning traces of varying lengths \citep{ma2025cot,munkhbat2025self}. While effective in shortening outputs, these methods demand substantial retraining cost and can distort the model’s output distribution, raising concerns about reliability and generalization to unseen tasks. Inference-time methods avoid retraining and instead attempt to stop reasoning dynamically by monitoring model signals such as logits \citep{yang2025dynamic} or intermediate answers \citep{fu2024efficiently}. Although these methods show that early stopping can reduce reasoning length without degrading accuracy, their reliance on probing introduces additional computation and often emphasizes token count reduction rather than true end-to-end latency improvements. 

\textbf{Speculative Decoding and Hidden States.}
Speculative decoding \citep{chen2023accelerating,leviathan2023fast} is a widely adopted technique for accelerating decoding speed, where a lightweight draft model proposes candidate tokens that a larger target model verifies in a single pass. Recent methods \citep{cai2024medusa, li2024eagle, li2024eagle2, li2025eagle3scalinginferenceacceleration, zhang_learning_2025} leverage hidden states to predict multiple future tokens. Beyond speculative decoding, several studies \citep{lin_controlling_2025, zhang_soft_2025, dong_emergent_2025, zhang_reasoning_2025}, have revealed that hidden states contain broader information about future outputs, including correctnes, response length, and reasoning paths. 
Building on this insight, our method extends speculative decoding by training hidden states not only to forecast future tokens but also to produce an early-exit signal.

\section{Conclusion}
\label{conclusion}
In this work, we propose \textbf{SpecExit}, a reasoning-aware early-exit framework that leverages latent signals from models' hidden states to dynamically terminate reasoning processes in LRMs. By concatenating auxiliary prediction heads to a lightweight draft model, SpecExit simultaneously predicts future tokens and early-exit signals in a single forward pass, eliminating the probing overhead required by previous approaches. Our experiments across diverse tasks and models demonstrate that SpecExit substantially reduces reasoning length by 66\% and achieves significant end-to-end latency improvements up to 2.5x without compromising accuracy. The proposed method highlights the potential of hidden states as informative signals for efficient reasoning and establishes a practical pathway for deploying LRMs in real-world scenarios. 

\bibliography{iclr2026_conference}

\begin{thebibliography}{35}
\providecommand{\natexlab}[1]{#1}
\providecommand{\url}[1]{\texttt{#1}}
\expandafter\ifx\csname urlstyle\endcsname\relax
  \providecommand{\doi}[1]{doi: #1}\else
  \providecommand{\doi}{doi: \begingroup \urlstyle{rm}\Url}\fi

\bibitem[Aggarwal \& Welleck(2025)Aggarwal and Welleck]{aggarwal2025l1}
Pranjal Aggarwal and Sean Welleck.
\newblock L1: Controlling how long a reasoning model thinks with reinforcement learning.
\newblock \emph{arXiv preprint arXiv:2503.04697}, 2025.

\bibitem[Brown et~al.()Brown, Juravsky, Ehrlich, Clark, Le, Ré, and Mirhoseini]{brown2024monkeys}
Bradley Brown, Jordan Juravsky, Ryan Ehrlich, Ronald Clark, Quoc~V. Le, Christopher Ré, and Azalia Mirhoseini.
\newblock Large language monkeys: Scaling inference compute with repeated sampling.
\newblock URL \url{http://arxiv.org/abs/2407.21787}.
\newblock version: 1.

\bibitem[Cai et~al.(2024)Cai, Li, Geng, Peng, Lee, Chen, and Dao]{cai2024medusa}
Tianle Cai, Yuhong Li, Zhengyang Geng, Hongwu Peng, Jason~D Lee, Deming Chen, and Tri Dao.
\newblock Medusa: Simple llm inference acceleration framework with multiple decoding heads.
\newblock \emph{arXiv preprint arXiv:2401.10774}, 2024.

\bibitem[Chen et~al.(2023)Chen, Borgeaud, Irving, Lespiau, Sifre, and Jumper]{chen2023accelerating}
Charlie Chen, Sebastian Borgeaud, Geoffrey Irving, Jean-Baptiste Lespiau, Laurent Sifre, and John Jumper.
\newblock Accelerating large language model decoding with speculative sampling.
\newblock \emph{arXiv preprint arXiv:2302.01318}, 2023.

\bibitem[Chen et~al.(2025)Chen, Xu, Liang, He, Pang, Yu, Song, Liu, Zhou, Zhang, Wang, Tu, Mi, and Yu]{chen2025think23overthinkingo1like}
Xingyu Chen, Jiahao Xu, Tian Liang, Zhiwei He, Jianhui Pang, Dian Yu, Linfeng Song, Qiuzhi Liu, Mengfei Zhou, Zhuosheng Zhang, Rui Wang, Zhaopeng Tu, Haitao Mi, and Dong Yu.
\newblock Do not think that much for 2+3=? on the overthinking of o1-like llms, 2025.
\newblock URL \url{https://arxiv.org/abs/2412.21187}.

\bibitem[Clark et~al.(2018)Clark, Cowhey, Etzioni, Khot, Sabharwal, Schoenick, and Tafjord]{clark2018think}
Peter Clark, Isaac Cowhey, Oren Etzioni, Tushar Khot, Ashish Sabharwal, Carissa Schoenick, and Oyvind Tafjord.
\newblock Think you have solved question answering? try arc, the ai2 reasoning challenge.
\newblock \emph{arXiv preprint arXiv:1803.05457}, 2018.

\bibitem[Cobbe et~al.(2021)Cobbe, Kosaraju, Bavarian, Chen, Jun, Kaiser, Plappert, Tworek, Hilton, Nakano, et~al.]{cobbe2021training}
Karl Cobbe, Vineet Kosaraju, Mohammad Bavarian, Mark Chen, Heewoo Jun, Lukasz Kaiser, Matthias Plappert, Jerry Tworek, Jacob Hilton, Reiichiro Nakano, et~al.
\newblock Training verifiers to solve math word problems.
\newblock \emph{arXiv preprint arXiv:2110.14168}, 2021.

\bibitem[DeepSeek-AI et~al.(2025)DeepSeek-AI, Guo, Yang, Zhang, Song, Zhang, Xu, Zhu, Ma, Wang, Bi, Zhang, Yu, Wu, Wu, Gou, Shao, Li, Gao, Liu, Xue, Wang, Wu, Feng, Lu, Zhao, Deng, Zhang, Ruan, Dai, Chen, Ji, Li, Lin, Dai, Luo, Hao, Chen, Li, Zhang, Bao, Xu, Wang, Ding, Xin, Gao, Qu, Li, Guo, Li, Wang, Chen, Yuan, Qiu, Li, Cai, Ni, Liang, Chen, Dong, Hu, Gao, Guan, Huang, Yu, Wang, Zhang, Zhao, Wang, Zhang, Xu, Xia, Zhang, Zhang, Tang, Li, Wang, Li, Tian, Huang, Zhang, Wang, Chen, Du, Ge, Zhang, Pan, Wang, Chen, Jin, Chen, Lu, Zhou, Chen, Ye, Wang, Yu, Zhou, Pan, Li, Zhou, Wu, Ye, Yun, Pei, Sun, Wang, Zeng, Zhao, Liu, Liang, Gao, Yu, Zhang, Xiao, An, Liu, Wang, Chen, Nie, Cheng, Liu, Xie, Liu, Yang, Li, Su, Lin, Li, Jin, Shen, Chen, Sun, Wang, Song, Zhou, Wang, Shan, Li, Wang, Wei, Zhang, Xu, Li, Zhao, Sun, Wang, Yu, Zhang, Shi, Xiong, He, Piao, Wang, Tan, Ma, Liu, Guo, Ou, Wang, Gong, Zou, He, Xiong, Luo, You, Liu, Zhou, Zhu, Xu, Huang, Li, Zheng, Zhu, Ma, Tang, Zha, Yan, Ren, Ren, Sha, Fu, Xu, Xie, Zhang,
  Hao, Ma, Yan, Wu, Gu, Zhu, Liu, Li, Xie, Song, Pan, Huang, Xu, Zhang, and Zhang]{deepseekai2025deepseekr1incentivizingreasoningcapability}
DeepSeek-AI, Daya Guo, Dejian Yang, Haowei Zhang, Junxiao Song, Ruoyu Zhang, Runxin Xu, Qihao Zhu, Shirong Ma, Peiyi Wang, Xiao Bi, Xiaokang Zhang, Xingkai Yu, Yu~Wu, Z.~F. Wu, Zhibin Gou, Zhihong Shao, Zhuoshu Li, Ziyi Gao, Aixin Liu, Bing Xue, Bingxuan Wang, Bochao Wu, Bei Feng, Chengda Lu, Chenggang Zhao, Chengqi Deng, Chenyu Zhang, Chong Ruan, Damai Dai, Deli Chen, Dongjie Ji, Erhang Li, Fangyun Lin, Fucong Dai, Fuli Luo, Guangbo Hao, Guanting Chen, Guowei Li, H.~Zhang, Han Bao, Hanwei Xu, Haocheng Wang, Honghui Ding, Huajian Xin, Huazuo Gao, Hui Qu, Hui Li, Jianzhong Guo, Jiashi Li, Jiawei Wang, Jingchang Chen, Jingyang Yuan, Junjie Qiu, Junlong Li, J.~L. Cai, Jiaqi Ni, Jian Liang, Jin Chen, Kai Dong, Kai Hu, Kaige Gao, Kang Guan, Kexin Huang, Kuai Yu, Lean Wang, Lecong Zhang, Liang Zhao, Litong Wang, Liyue Zhang, Lei Xu, Leyi Xia, Mingchuan Zhang, Minghua Zhang, Minghui Tang, Meng Li, Miaojun Wang, Mingming Li, Ning Tian, Panpan Huang, Peng Zhang, Qiancheng Wang, Qinyu Chen, Qiushi Du, Ruiqi Ge, Ruisong
  Zhang, Ruizhe Pan, Runji Wang, R.~J. Chen, R.~L. Jin, Ruyi Chen, Shanghao Lu, Shangyan Zhou, Shanhuang Chen, Shengfeng Ye, Shiyu Wang, Shuiping Yu, Shunfeng Zhou, Shuting Pan, S.~S. Li, Shuang Zhou, Shaoqing Wu, Shengfeng Ye, Tao Yun, Tian Pei, Tianyu Sun, T.~Wang, Wangding Zeng, Wanjia Zhao, Wen Liu, Wenfeng Liang, Wenjun Gao, Wenqin Yu, Wentao Zhang, W.~L. Xiao, Wei An, Xiaodong Liu, Xiaohan Wang, Xiaokang Chen, Xiaotao Nie, Xin Cheng, Xin Liu, Xin Xie, Xingchao Liu, Xinyu Yang, Xinyuan Li, Xuecheng Su, Xuheng Lin, X.~Q. Li, Xiangyue Jin, Xiaojin Shen, Xiaosha Chen, Xiaowen Sun, Xiaoxiang Wang, Xinnan Song, Xinyi Zhou, Xianzu Wang, Xinxia Shan, Y.~K. Li, Y.~Q. Wang, Y.~X. Wei, Yang Zhang, Yanhong Xu, Yao Li, Yao Zhao, Yaofeng Sun, Yaohui Wang, Yi~Yu, Yichao Zhang, Yifan Shi, Yiliang Xiong, Ying He, Yishi Piao, Yisong Wang, Yixuan Tan, Yiyang Ma, Yiyuan Liu, Yongqiang Guo, Yuan Ou, Yuduan Wang, Yue Gong, Yuheng Zou, Yujia He, Yunfan Xiong, Yuxiang Luo, Yuxiang You, Yuxuan Liu, Yuyang Zhou, Y.~X. Zhu,
  Yanhong Xu, Yanping Huang, Yaohui Li, Yi~Zheng, Yuchen Zhu, Yunxian Ma, Ying Tang, Yukun Zha, Yuting Yan, Z.~Z. Ren, Zehui Ren, Zhangli Sha, Zhe Fu, Zhean Xu, Zhenda Xie, Zhengyan Zhang, Zhewen Hao, Zhicheng Ma, Zhigang Yan, Zhiyu Wu, Zihui Gu, Zijia Zhu, Zijun Liu, Zilin Li, Ziwei Xie, Ziyang Song, Zizheng Pan, Zhen Huang, Zhipeng Xu, Zhongyu Zhang, and Zhen Zhang.
\newblock Deepseek-r1: Incentivizing reasoning capability in llms via reinforcement learning, 2025.
\newblock URL \url{https://arxiv.org/abs/2501.12948}.

\bibitem[Dong et~al.()Dong, Zhou, Liu, Yang, and Lu]{dong_emergent_2025}
Zhichen Dong, Zhanhui Zhou, Zhixuan Liu, Chao Yang, and Chaochao Lu.
\newblock Emergent response planning in {LLMs}.
\newblock URL \url{http://arxiv.org/abs/2502.06258}.

\bibitem[Fu et~al.(2024)Fu, Chen, Zhu, Fu, Dai, Qiao, and Zhang]{fu2024efficiently}
Yichao Fu, Junda Chen, Siqi Zhu, Zheyu Fu, Zhongdongming Dai, Aurick Qiao, and Hao Zhang.
\newblock Efficiently serving llm reasoning programs with certaindex.
\newblock \emph{arXiv e-prints}, pp.\  arXiv--2412, 2024.

\bibitem[Hendrycks et~al.(2021)Hendrycks, Burns, Kadavath, Arora, Basart, Tang, Song, and Steinhardt]{math500hendrycks2021measuringmathematicalproblemsolving}
Dan Hendrycks, Collin Burns, Saurav Kadavath, Akul Arora, Steven Basart, Eric Tang, Dawn Song, and Jacob Steinhardt.
\newblock Measuring mathematical problem solving with the math dataset, 2021.
\newblock URL \url{https://arxiv.org/abs/2103.03874}.

\bibitem[Kwon et~al.(2023)Kwon, Li, Zhuang, Sheng, Zheng, Yu, Gonzalez, Zhang, and Stoica]{woosuk2023vllm}
Woosuk Kwon, Zhuohan Li, Siyuan Zhuang, Ying Sheng, Lianmin Zheng, Cody~Hao Yu, Joseph Gonzalez, Hao Zhang, and Ion Stoica.
\newblock Efficient memory management for large language model serving with pagedattention.
\newblock In \emph{Proceedings of the 29th Symposium on Operating Systems Principles}, SOSP '23, pp.\  611–626, New York, NY, USA, 2023. Association for Computing Machinery.
\newblock ISBN 9798400702297.
\newblock \doi{10.1145/3600006.3613165}.
\newblock URL \url{https://doi.org/10.1145/3600006.3613165}.

\bibitem[Leviathan et~al.(2023)Leviathan, Kalman, and Matias]{leviathan2023fast}
Yaniv Leviathan, Matan Kalman, and Yossi Matias.
\newblock Fast inference from transformers via speculative decoding.
\newblock In \emph{International Conference on Machine Learning}, pp.\  19274--19286. PMLR, 2023.

\bibitem[Li et~al.(2024{\natexlab{a}})Li, Wei, Zhang, and Zhang]{li2024eagle}
Yuhui Li, Fangyun Wei, Chao Zhang, and Hongyang Zhang.
\newblock {EAGLE}: Speculative sampling requires rethinking feature uncertainty.
\newblock In \emph{International Conference on Machine Learning}, 2024{\natexlab{a}}.

\bibitem[Li et~al.(2024{\natexlab{b}})Li, Wei, Zhang, and Zhang]{li2024eagle2}
Yuhui Li, Fangyun Wei, Chao Zhang, and Hongyang Zhang.
\newblock {EAGLE-2}: Faster inference of language models with dynamic draft trees.
\newblock In \emph{Empirical Methods in Natural Language Processing}, 2024{\natexlab{b}}.

\bibitem[Li et~al.(2025)Li, Wei, Zhang, and Zhang]{li2025eagle3scalinginferenceacceleration}
Yuhui Li, Fangyun Wei, Chao Zhang, and Hongyang Zhang.
\newblock {EAGLE-3}: Scaling up inference acceleration of large language models via training-time test, 2025.
\newblock URL \url{https://arxiv.org/abs/2503.01840}.

\bibitem[Lin et~al.()Lin, Fu, Chen, Chen, Xie, Wang, Cai, Wang, and Ye]{lin_controlling_2025}
Zhengkai Lin, Zhihang Fu, Ze~Chen, Chao Chen, Liang Xie, Wenxiao Wang, Deng Cai, Zheng Wang, and Jieping Ye.
\newblock Controlling thinking speed in reasoning models.
\newblock URL \url{http://arxiv.org/abs/2507.03704}.
\newblock version: 1.

\bibitem[Liu et~al.(2023)Liu, Xia, Wang, and Zhang]{liu2023your}
Jiawei Liu, Chunqiu~Steven Xia, Yuyao Wang, and Lingming Zhang.
\newblock Is your code generated by chatgpt really correct? rigorous evaluation of large language models for code generation.
\newblock \emph{Advances in Neural Information Processing Systems}, 36:\penalty0 21558--21572, 2023.

\bibitem[Ma et~al.(2025{\natexlab{a}})Ma, He, Snell, Griggs, Min, and Zaharia]{ma2025reasoning}
Wenjie Ma, Jingxuan He, Charlie Snell, Tyler Griggs, Sewon Min, and Matei Zaharia.
\newblock Reasoning models can be effective without thinking.
\newblock \emph{arXiv preprint arXiv:2504.09858}, 2025{\natexlab{a}}.

\bibitem[Ma et~al.(2025{\natexlab{b}})Ma, Wan, Yu, Fang, and Wang]{ma2025cot}
Xinyin Ma, Guangnian Wan, Runpeng Yu, Gongfan Fang, and Xinchao Wang.
\newblock Cot-valve: Length-compressible chain-of-thought tuning.
\newblock \emph{arXiv preprint arXiv:2502.09601}, 2025{\natexlab{b}}.

\bibitem[\mbox{MAA Committees}()]{aime}
\mbox{MAA Committees}.
\newblock Aime problems and solutions.
\newblock \url{https://artofproblemsolving.com/wiki/index.php/AIME_Problems_and_Solutions}.

\bibitem[Muennighoff et~al.()Muennighoff, Yang, Shi, Li, Fei-Fei, Hajishirzi, Zettlemoyer, Liang, Candès, and Hashimoto]{muennighoff2025s1}
Niklas Muennighoff, Zitong Yang, Weijia Shi, Xiang~Lisa Li, Li~Fei-Fei, Hannaneh Hajishirzi, Luke Zettlemoyer, Percy Liang, Emmanuel Candès, and Tatsunori Hashimoto.
\newblock S1: Simple test-time scaling.
\newblock URL \url{http://arxiv.org/abs/2501.19393}.
\newblock version: 1.

\bibitem[Munkhbat et~al.(2025)Munkhbat, Ho, Kim, Yang, Kim, and Yun]{munkhbat2025self}
Tergel Munkhbat, Namgyu Ho, Seo~Hyun Kim, Yongjin Yang, Yujin Kim, and Se-Young Yun.
\newblock Self-training elicits concise reasoning in large language models.
\newblock \emph{arXiv preprint arXiv:2502.20122}, 2025.

\bibitem[OpenAI(2024)]{o1}
OpenAI.
\newblock Learning to reason with llms, September 2024.
\newblock URL \url{https://openai.com/index/learning-to-reason-with-llms/}.

\bibitem[Qwen et~al.(2025)Qwen, :, Yang, Yang, Zhang, Hui, Zheng, Yu, Li, Liu, Huang, Wei, Lin, Yang, Tu, Zhang, Yang, Yang, Zhou, Lin, Dang, Lu, Bao, Yang, Yu, Li, Xue, Zhang, Zhu, Men, Lin, Li, Tang, Xia, Ren, Ren, Fan, Su, Zhang, Wan, Liu, Cui, Zhang, and Qiu]{qwen2025qwen25technicalreport}
Qwen, :, An~Yang, Baosong Yang, Beichen Zhang, Binyuan Hui, Bo~Zheng, Bowen Yu, Chengyuan Li, Dayiheng Liu, Fei Huang, Haoran Wei, Huan Lin, Jian Yang, Jianhong Tu, Jianwei Zhang, Jianxin Yang, Jiaxi Yang, Jingren Zhou, Junyang Lin, Kai Dang, Keming Lu, Keqin Bao, Kexin Yang, Le~Yu, Mei Li, Mingfeng Xue, Pei Zhang, Qin Zhu, Rui Men, Runji Lin, Tianhao Li, Tianyi Tang, Tingyu Xia, Xingzhang Ren, Xuancheng Ren, Yang Fan, Yang Su, Yichang Zhang, Yu~Wan, Yuqiong Liu, Zeyu Cui, Zhenru Zhang, and Zihan Qiu.
\newblock Qwen2.5 technical report, 2025.
\newblock URL \url{https://arxiv.org/abs/2412.15115}.

\bibitem[Rein et~al.(2023)Rein, Hou, Stickland, Petty, Pang, Dirani, Michael, and Bowman]{rein2023gpqagraduatelevelgoogleproofqa}
David Rein, Betty~Li Hou, Asa~Cooper Stickland, Jackson Petty, Richard~Yuanzhe Pang, Julien Dirani, Julian Michael, and Samuel~R. Bowman.
\newblock Gpqa: A graduate-level google-proof q\&a benchmark, 2023.
\newblock URL \url{https://arxiv.org/abs/2311.12022}.

\bibitem[Shen et~al.(2025)Shen, Zhang, Huang, Shi, Zhang, Yan, Wang, Wang, and Lian]{shen2025dast}
Yi~Shen, Jian Zhang, Jieyun Huang, Shuming Shi, Wenjing Zhang, Jiangze Yan, Ning Wang, Kai Wang, and Shiguo Lian.
\newblock Dast: Difficulty-adaptive slow-thinking for large reasoning models.
\newblock \emph{arXiv preprint arXiv:2503.04472}, 2025.

\bibitem[Snell et~al.()Snell, Lee, Xu, and Kumar]{snell2024scaling}
Charlie Snell, Jaehoon Lee, Kelvin Xu, and Aviral Kumar.
\newblock Scaling llm test-time compute optimally can be more effective than scaling model parameters.
\newblock URL \url{http://arxiv.org/abs/2408.03314}.
\newblock version: 1.

\bibitem[Sui et~al.(2025)Sui, Chuang, Wang, Zhang, Zhang, Yuan, Liu, Wen, Zhong, Zou, Chen, and Hu]{sui2025stopoverthinkingsurveyefficient}
Yang Sui, Yu-Neng Chuang, Guanchu Wang, Jiamu Zhang, Tianyi Zhang, Jiayi Yuan, Hongyi Liu, Andrew Wen, Shaochen Zhong, Na~Zou, Hanjie Chen, and Xia Hu.
\newblock Stop overthinking: A survey on efficient reasoning for large language models, 2025.
\newblock URL \url{https://arxiv.org/abs/2503.16419}.

\bibitem[Wei et~al.()Wei, Wang, Schuurmans, Bosma, Xia, Chi, Le, and Zhou]{wei2022chain}
Jason Wei, Xuezhi Wang, Dale Schuurmans, Maarten Bosma, Fei Xia, Ed~Chi, Quoc~V. Le, and Denny Zhou.
\newblock Chain-of-thought prompting elicits reasoning in large language models.
\newblock URL \url{http://arxiv.org/abs/2201.11903}.
\newblock version: 1.

\bibitem[Yang et~al.(2025)Yang, Si, Duan, Zhu, Zhu, Lin, Cao, and Wang]{yang2025dynamic}
Chenxu Yang, Qingyi Si, Yongjie Duan, Zheliang Zhu, Chenyu Zhu, Zheng Lin, Li~Cao, and Weiping Wang.
\newblock Dynamic early exit in reasoning models.
\newblock \emph{arXiv preprint arXiv:2504.15895}, 2025.

\bibitem[Yeo et~al.(2025)Yeo, Tong, Niu, Neubig, and Yue]{yeo2025demystifying}
Edward Yeo, Yuxuan Tong, Morry Niu, Graham Neubig, and Xiang Yue.
\newblock Demystifying long chain-of-thought reasoning in llms.
\newblock \emph{arXiv preprint arXiv:2502.03373}, 2025.

\bibitem[Zhang et~al.({\natexlab{a}})Zhang, Chen, Pan, Zhao, Panda, Li, and He]{zhang_reasoning_2025}
Anqi Zhang, Yulin Chen, Jane Pan, Chen Zhao, Aurojit Panda, Jinyang Li, and He~He.
\newblock Reasoning models know when they’re right: Probing hid- den states for self-verification.
\newblock {\natexlab{a}}.

\bibitem[Zhang et~al.({\natexlab{b}})Zhang, Wang, Huang, and Xu]{zhang_learning_2025}
Lefan Zhang, Xiaodan Wang, Yanhua Huang, and Ruiwen Xu.
\newblock Learning harmonized representations for speculative sampling, {\natexlab{b}}.
\newblock URL \url{http://arxiv.org/abs/2408.15766}.

\bibitem[Zhang et~al.({\natexlab{c}})Zhang, He, Yan, Shen, Zhao, Wang, Shen, and Wang]{zhang_soft_2025}
Zhen Zhang, Xuehai He, Weixiang Yan, Ao~Shen, Chenyang Zhao, Shuohang Wang, Yelong Shen, and Xin~Eric Wang.
\newblock Soft thinking: Unlocking the reasoning potential of {LLMs} in continuous concept space, {\natexlab{c}}.
\newblock URL \url{http://arxiv.org/abs/2505.15778}.

\end{thebibliography}
\bibliographystyle{iclr2026_conference}

\appendix
\newpage
\section{Appendix}
This section may contain supplementary materials such as additional experimental details, ablation studies, hyperparameter settings, and qualitative examples of generated CoT sequences.


\subsection{Implementation details}

\textbf{Stop Threshold}: In the ablation study of SpecExit signal types, the following thresholds are applied as stopping conditions for the respective signal types: SpecExit-Confidence requires predicted confidence value greater than 0.9, SpecExit-Progress requires predicted progress value greater than 0.8, SpecExit-Remain requires predicted remaining reasoning length value less than 100, and SpecExit* combines thresholds with a predicted confidence value greater than 0.8, predicted progress greater than 0.3, and predicted remaining reasoning length less than 200.

\textbf{Signal Smoothing}: In the ablation study of signal smoothing strategies, the following methods are implemented to stabilize cognitive signals for early-exit decisions:

\begin{itemize}[leftmargin=*,itemsep=0.15em]
    \item Sliding Window: The sliding window approach smooths the signal by averaging the last $N$ predicted signal values, with $N$ set to 10. The mean score $x_t$ at decoding step $t$ is computed as:
    \begin{equation}
       x_t = \text{Mean}(s_t, N) = \frac{1}{N} \sum_{i=t-N+1}^{t} s_i,
    \end{equation}
    where $s_i$ denotes the predicted signal value at decoding step $i$.

    \item Momentum-based Prediction: This method predicts the next score based on the momentum, which is calculated as the difference between $N$ consecutive signal values, with $N$ set to 10. The predicted score $x_t$ at decoding step $t$ is given by:
    \begin{equation}
       x_t = \text{Predict}(s_t, N) = s_{t-1} + \frac{1}{N-1} \sum_{i=t-N+1}^{t-1} (s_i - s_{i-1}).
    \end{equation}

     \item Paragraph Mean: In this approach, the score $x_t$ is calculated as the average of all predicted signal values within the current paragraph:
    \begin{equation}
       x_t = \frac{1}{T} \sum_{i=1}^{T} s_i,
    \end{equation}
    where $T$ is the total number of steps in the current paragraph.

    \item Exponential Weighted Moving Average (EWMA): In this approach, the smoothing factor $\alpha$ is set to 0.1. The new score $x_t$ is updated based on the previous score $x_{t-1}$ and the current signal value $s_t$ as:
    \begin{equation}
       x_t = \text{EWMA}(s_t, x_{t-1}, \alpha) = \alpha \cdot s_t + (1 - \alpha) \cdot x_{t-1}.
    \end{equation}
\end{itemize}

\begin{figure}[!hbp]
\begin{center}
\includegraphics[width=0.65\linewidth]{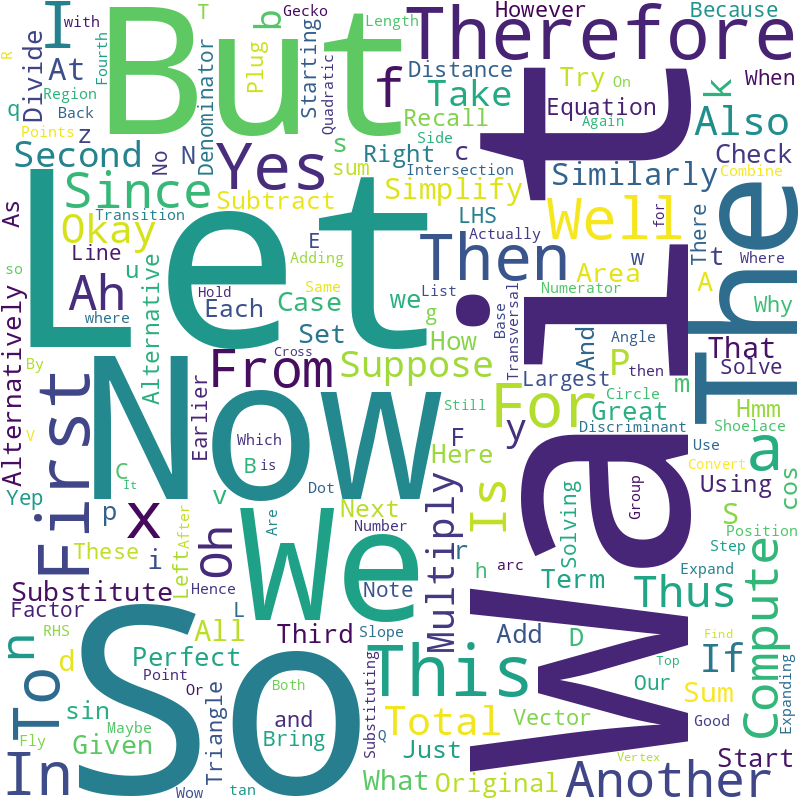}
\end{center}
\caption{Discourse marker distribution in Qwen3-4B-Thinking-2507’s responses on the MATH500 \citep{math500hendrycks2021measuringmathematicalproblemsolving} dataset.}
\label{fig:wordcloud}
\end{figure}

\textbf{Discourse markers}:
We collect high-frequency words appearing at the beginning of model-generated sentences as discourse markers. As shown in Figure~\ref{fig:wordcloud}, we present examples of discourse markers extracted from Qwen3-4B-Thinking-2507 on the MATH dataset. Among them, transitional words such as "Wait" and "But" can be regarded as a subset of these high-frequency markers.

\subsection{Predicted Signal Visualization}

\begin{figure*}[htbp]
    \centering
    \begin{subfigure}[t]{0.93\textwidth}
        \centering
        \includegraphics[width=\textwidth]{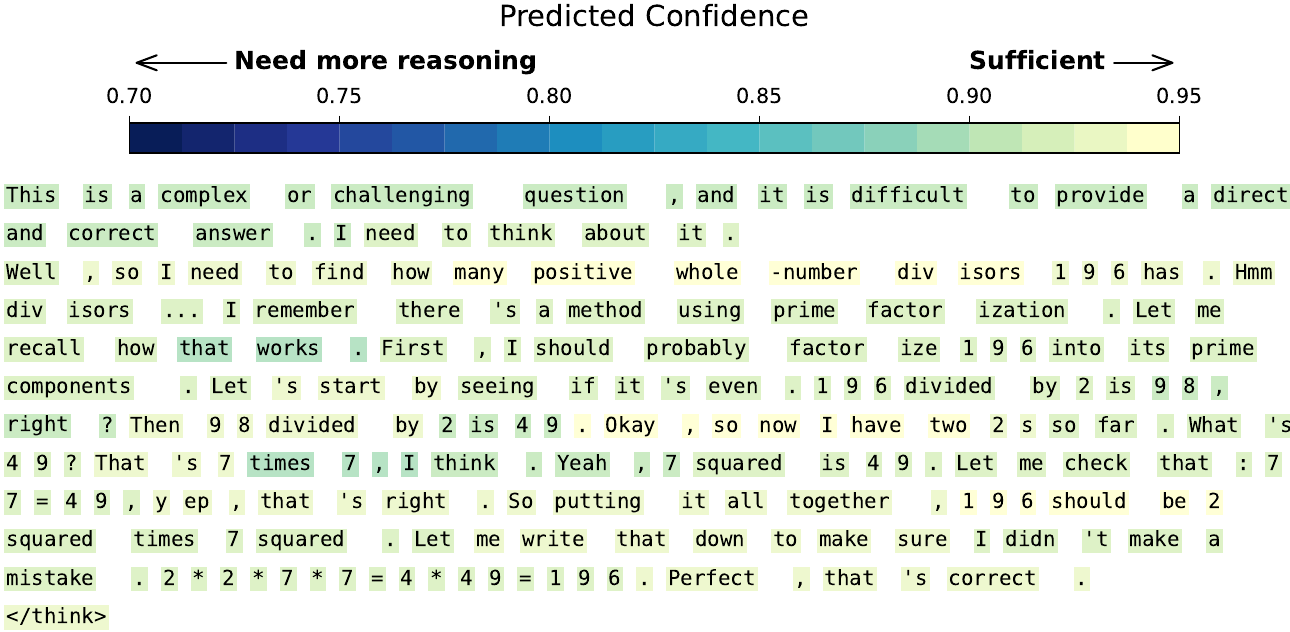}
        \label{fig:math_pred_confidence}
    \end{subfigure}


    \begin{subfigure}[t]{0.93\textwidth}
        \centering
        \includegraphics[width=\textwidth]{figs/signals/math/pred_progress.pdf}
        \label{fig:math_pred_progress}
    \end{subfigure}


    \begin{subfigure}[t]{0.93\textwidth}
        \centering
        \includegraphics[width=\textwidth]{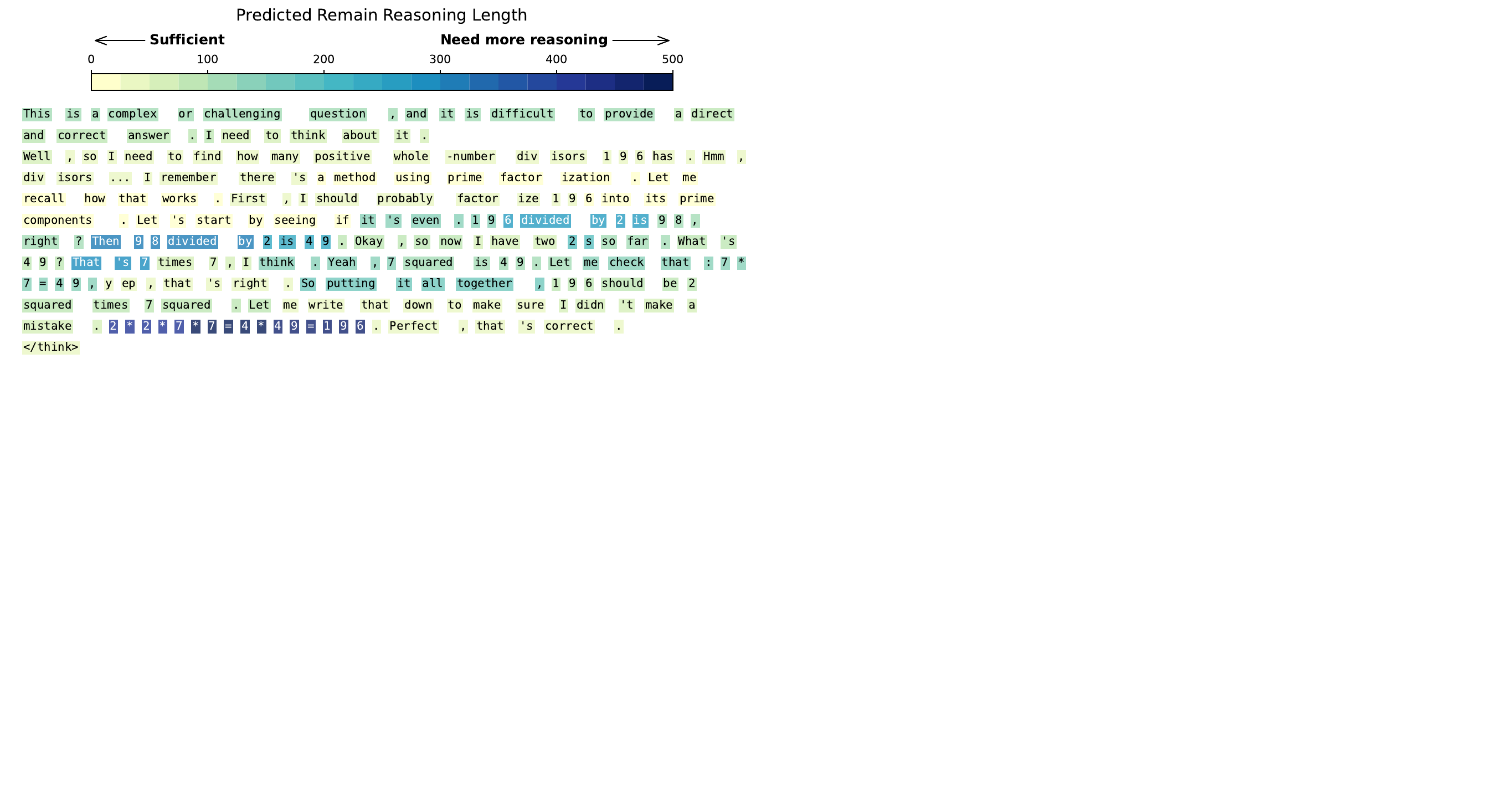}
        \label{fig:math_pred_remain}
    \end{subfigure}

    \caption{Visualization of reasoning signals for a simple problem, illustrated with an example from the MATH500 \citep{math500hendrycks2021measuringmathematicalproblemsolving} dataset, where darker colors denote insufficient reasoning and lighter colors denote sufficiency.}
    \label{fig:math_signals}
\end{figure*}

\begin{figure*}[htbp]
    \centering
    \begin{subfigure}[t]{0.62\textwidth}
        \centering
        \includegraphics[width=\textwidth]{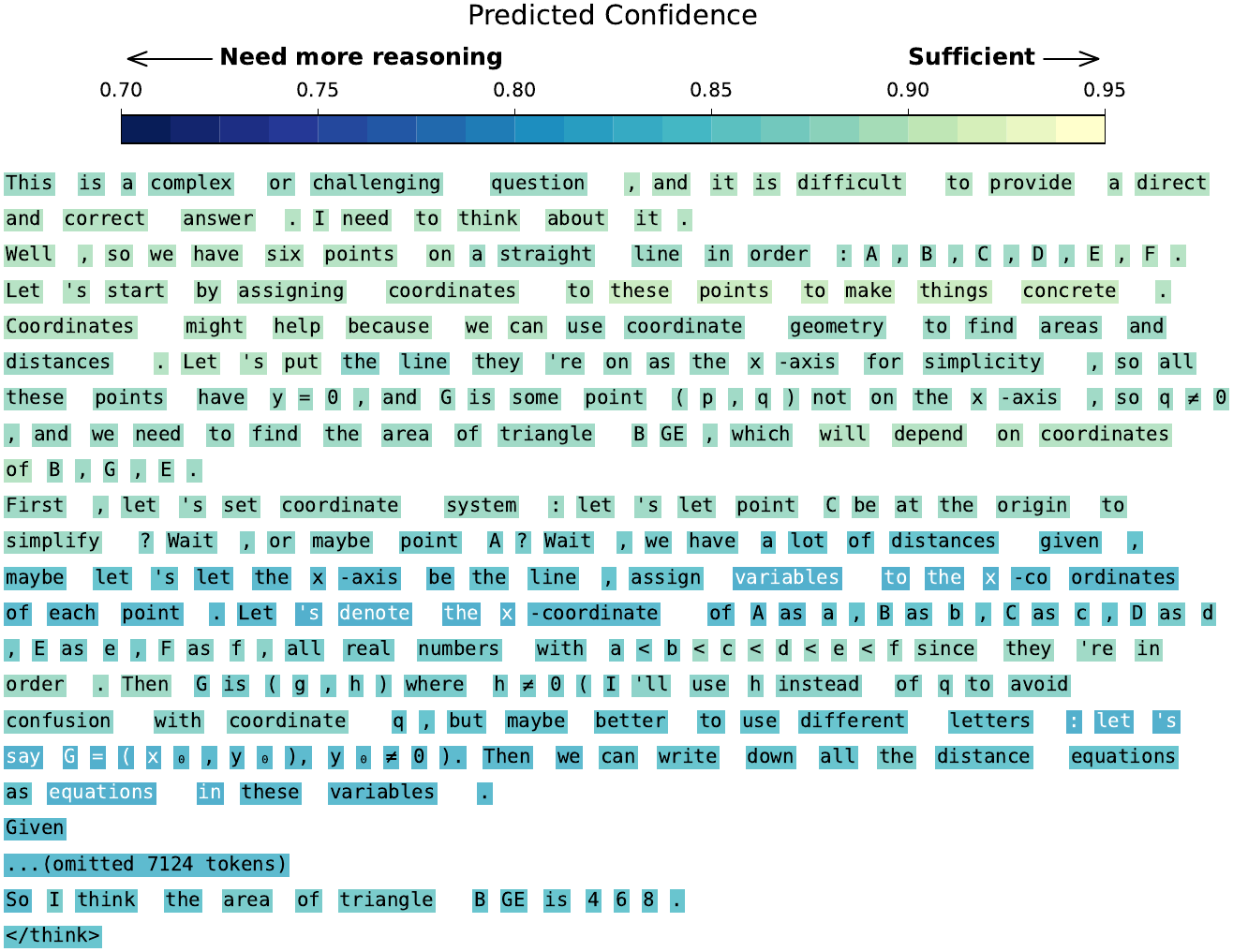}
        \label{fig:aime_pred_confidence}
    \end{subfigure}


    \begin{subfigure}[t]{0.62\textwidth}
        \centering
        \includegraphics[width=\textwidth]{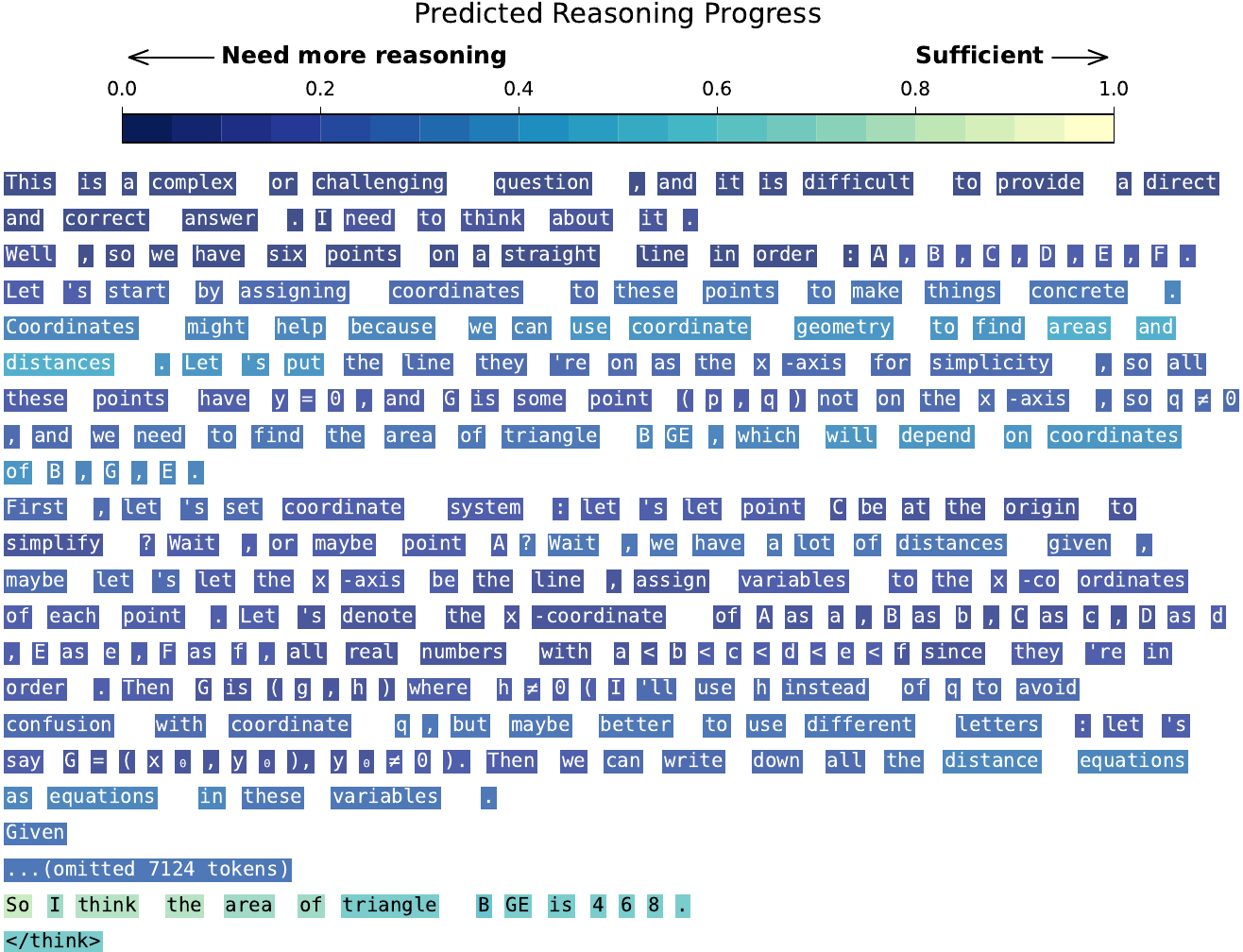}
        \label{fig:aime_pred_progress}
    \end{subfigure}


    \begin{subfigure}[t]{0.62\textwidth}
        \centering
        \includegraphics[width=\textwidth]{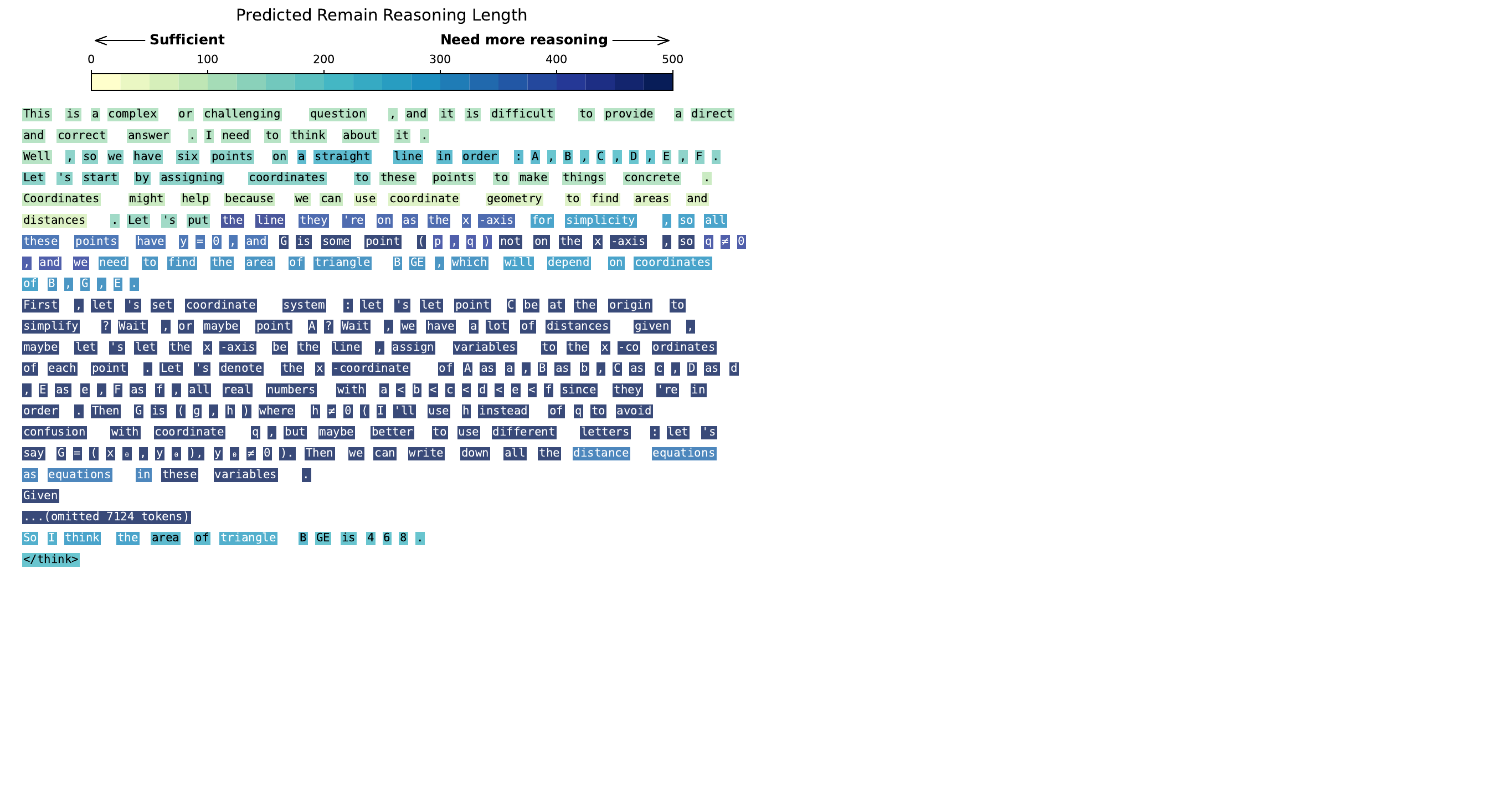}
        \label{fig:aime_pred_remain}
    \end{subfigure}

    \caption{Visualization of reasoning signals for a complicated problem, illustrated with an example from the AIME \citep{aime}  dataset, where darker colors denote insufficient reasoning and lighter colors denote sufficiency.}
    \label{fig:aime_signals}
\end{figure*}

As shown in Figure~\ref{fig:math_signals}, for simple problems, the predicted confidence remains high, the predicted remaining reasoning length is relatively short, and the predicted progress rises rapidly within the first few sentences, with only minor drops on a few uncertain words. This indicates that for such problems, the model is able to establish a stable reasoning trajectory at an early stage. In contrast, as shown in Figure~\ref{fig:aime_signals}, for complicated problems, the model also exhibits high confidence and a short remaining reasoning length in the initial summarization phase, but once it enters the detailed analysis stage, the predicted remaining reasoning length increases significantly, confidence drops, and progress, though rising in the beginning, fluctuates heavily during repeated self-reflection, making it difficult to stabilize at a high threshold.

These observations reveal the inherent limitations of relying on individual signals for early exiting. When depending solely on confidence, the model often exhibits overconfidence and terminates too early before sufficient reasoning has been completed, leading to substantial accuracy degradation. When depending solely on the predicted remaining reasoning length, the model may become overly optimistic in the early stages of complicated problems, resulting in premature exits before essential reasoning steps are accomplished. When depending solely on progress, the signal tends to fluctuate and remain unstable in complex reasoning tasks, making it difficult to trigger an appropriate early exit and thereby restricting achievable speedup. In summary, each single signal suffers from the inability to balance accuracy and efficiency across diverse problem types. By integrating multiple signals in a complementary manner, the model can achieve a smoother trade-off between reasoning accuracy and inference acceleration.

\subsection{Case Study Details}

\begin{figure}[h]
\begin{center}
\includegraphics[width=\linewidth]{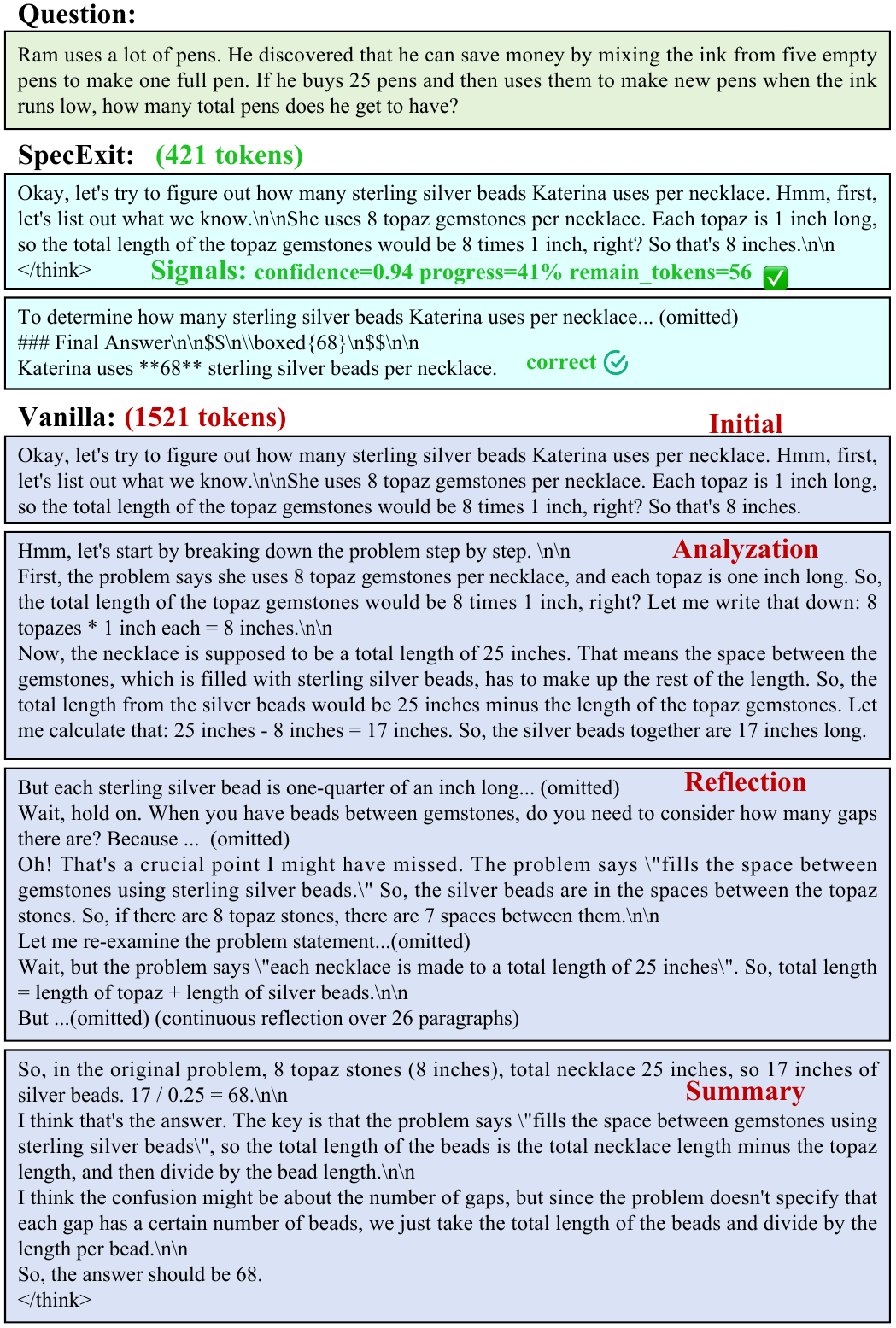}
\end{center}
\caption{Discourse marker distribution in Qwen3-4B-Thinking-2507’s responses on the MATH500 \citep{math500hendrycks2021measuringmathematicalproblemsolving} dataset.}
\label{fig:case_study1}
\end{figure}

Figure~\ref{fig:case_study1} presents an example from the GSM8K \citep{cobbe2021training} dataset, where SpecExit is applied after an initial analysis. In this case, SpecExit inserts a decision to exit reasoning based on the signal magnitude after completing the first paragraph, thus preventing the continuation of redundant reasoning tokens. The process begins with the model evaluating the initial segment of the problem, analyzing the available context and producing intermediate reasoning steps. When SpecExit detects that all the signals exceeds a predefined threshold, it triggers the early exit mechanism. As a result, the model halts further reasoning and transitions directly to the answering phase, thus producing the correct result in a more efficient manner.

By shortening the unnecessary token generation and reducing redundant reasoning, SpecExit leads to significant performance improvements in terms of inference speed while maintaining the correctness of the answer. This early exit strategy not only accelerates the inference process but also avoids the potential confusion that the model might experience from excessive reflection or unnecessary computations. This case highlights the effectiveness of integrating early exit mechanisms in enhancing both the efficiency and reliability of large language model inference.

\end{document}